\documentclass[review]{elsarticle}

\usepackage{lineno,hyperref}

\usepackage{graphicx}
\usepackage{xcolor}
\usepackage{natbib}
\usepackage{wrapfig}
\usepackage[english]{babel}
\usepackage[center]{caption}

\usepackage{mathtools}
\usepackage{mathptmx}
\usepackage{amsmath}
\usepackage{amssymb}
\usepackage{latexsym}
\usepackage{amsfonts}

\usepackage[ruled, linesnumbered]{algorithm2e}
\usepackage{amsmath}

\usepackage[normalem]{ulem}
\useunder{\uline}{\ul}{}

\usepackage{capt-of, blindtext}
\usepackage{multirow}
\usepackage{multicol}
\usepackage{footnote}
\usepackage{url}
\usepackage{tikz}

\usepackage{pdfpages}
\usepackage{setspace}
\usepackage{lineno}
\usepackage{fullpage}
\usepackage{kpfonts}
\usepackage{setspace}
\usepackage{mdframed}
\usepackage{bbold}

\usepackage{enumitem}

\usepackage[bottom]{footmisc}

\makesavenoteenv{tabular}
\makesavenoteenv{table}

\newtheorem{myDef}{Definition}
\newcommand{\SET}[1]{\mathcal{#1}}

\newcommand{\s}[1]{\mathit{#1}}

\newcommand{\vect}[1]{\boldsymbol{#1}}

\mdfdefinestyle{MyFrame}{%
                linecolor=black,
                outerlinewidth=2pt,
                roundcorner=10pt,
                innertopmargin=0.3cm,
                innerbottommargin=0.3cm,
                innerrightmargin=10pt,
                innerleftmargin=10pt,
                backgroundcolor=white}

\modulolinenumbers[5]

\journal{arXiv}

\date{}

\newcommand{\highlight}[1]{\textcolor{black}{#1}}
\newcommand{\revision}[1]{\textcolor{black}{#1}}

\begin{document}

\begin{frontmatter}

\title{Building Interpretable Models for Business Process Prediction using Shared and Specialised Attention Mechanisms}

\author[addr1]{Bemali Wickramanayake}
\ead{bemali.wickramanayake@hdr.qut.edu.au}
\author[addr1]{Zhipeng He}
\ead{zhipeng.he@connect.qut.edu.au}
\author[addr1]{Chun Ouyang}
\ead{c.ouyang@qut.edu.au}
\author[addr1]{Catarina Moreira}
\ead{catarina.pintomoreira@qut.edu.au}
\author[addr1]{Yue Xu}
\ead{yue.xu@qut.edu.au}
\author[addr2]{Renuka Sindhgatta}
\ead{renuka.sr@ibm.com} 

\address[addr1]{Queensland University of Technology, Brisbane, Australia}
\address[addr2]{IBM Research, Bangalore, India}


\begin{abstract}


Predictive process analytics, often underpinned by deep learning techniques, is a newly emerged discipline dedicated for providing business process intelligence in modern organisations. Whilst accuracy has been a dominant criterion in building predictive capabilities, the use of deep learning techniques comes at the cost of the resulting models being used as `black boxes', i.e., they are unable to provide insights into why a certain business process prediction was made. So far, little attention has been paid to interpretability in the design of deep learning-based process predictive models. 

In this paper, we address the `black-box' problem in the context of predictive process analytics by developing attention-based models that are capable to inform both \textit{what} and \textit{why} is a process prediction. We propose i) two types of attentions --- \textit{event attention} to capture the impact of specific events on a prediction, and \textit{attribute attention} to reveal which attribute(s) of an event influenced the prediction; and ii) two attention mechanisms --- \textit{shared attention mechanism} and \textit{specialised attention mechanism} to reflect different design decisions between whether to construct attribute attention on individual input features (specialised) or using the concatenated feature tensor of all input feature vectors (shared). 
These lead to two distinct attention-based models, and both are interpretable models that incorporate interpretability directly into the structure of a process predictive model. We conduct experimental evaluation of the proposed models using real-life dataset and comparative analysis between the models for accuracy and interpretability, and draw insights from the evaluation and analysis results.

\revision{The results demonstrate that i) the proposed attention-based models can achieve reasonably high accuracy; ii) both are capable of providing relevant interpretations (when validated against domain knowledge); and iii) whilst the two models perform equally in terms of prediction accuracy, the specialised attention-based model tends to provide more relevant interpretations than the shared attention-based model, reflecting the fact that the specialised attention-based model is designed to facilitate better interpretability.} 
\end{abstract}


\end{frontmatter}


\section{Introduction}
\label{sec:intro}

%
%
%

Deep learning techniques have achieved a remarkable performance across a wide range of applications. However, they are also typical examples of `black-box' models due to their sophisticated internal representations. Consequently, despite the ever-increasing accuracy of deep learning models, the full adoption of deep learning-based solutions in real practice is being impeded by the lack of model transparency and explainability. The recent body of literature, known as explainable AI (XAI), has emphasised the need to improve the understanding and explainability of AI learning techniques such as deep neural networks, and proposed methods and techniques for explaining `black-box' models. But there is no `silver bullet' approach since the applications in which a `black-box' model may be used are diverse and each existing approach is typically developed to provide a solution for a specific problem~\cite{guidotti2018}.

Predictive process analytics is a relatively new branch of data analytics dedicated to providing business process intelligence in modern organisations. It uses event logs, which capture process execution traces in the form of multi-dimensional sequence data, as the key input to train predictive models. These predictive models, underpinned by advanced machine learning or deep learning techniques, can be used to make predictions about the future states of business process execution. In particular, recurrent neural networks (RNNs) and their variants such as long short term memory (LSTM) networks have naturally found applicability in predictive process analytics. For example, given an input event log which records the sequence data of a running business process, an LSTM-based model can be trained to predict the next event in the running process~\cite{Evermann2017,Tax2017,Camargo2019} as well as the remaining execution time of the running process~\cite{Camargo2019,verenich2019survey}.


Whilst accuracy has been a dominant criterion in building predictive capabilities in process analytics, the use of deep learning techniques as the underlying mechanism comes at the cost of the models being used as `black boxes' --- they are unable to provide insights into why a certain business process prediction was made. Recent studies have attempted to apply existing XAI methods to explain opaque process predictive models (e.g.,~\cite{Galanti2020,Sindhgatta2020b}). However, model explainability in the context of deep learning-based predictive process analytics is yet underexplored, and little attention has been paid to interpretability in the design of deep learning-based process predictive models. 

We aim to address the above gap in our work by building interpretable models for business process prediction using \textit{attention mechanisms}. In general, attention mechanisms allow a predictive model to focus on specific features in a given input sequence for a prediction task, and are often used to improve the accuracy of a predictive model. But they can also be applied to support model interpretability. Since attention-based models require the computation of a distribution of weights over inputs, the attention weights can provide some insights to a decision-maker of \textit{what} features influenced the prediction by manipulating the distribution of weights associated to the input~\cite{serrano2019attention}, and consequently these can help with understanding of \textit{why} a certain prediction was computed. 

In this paper, we put forward two types of attentions: \textit{event attention} and \textit{attribute attention}, as inspired by the work proposed in~\cite{Retain_Choi2016}. Given a running process execution trace, when predicting the activity of the next event in the trace, our model can provide intuitive information on specific events in the trace that influenced the prediction, by computing \textit{event attention}. Also, our model can reveal which attribute(s) of an event (e.g., activity, resource, or time) influenced the prediction, by computing \textit{attribute attention}. 

We compare two different attention mechanisms: \textit{shared attention mechanism}~\cite{Sindhgatta2020a} and \textit{specialised attention mechanism} proposed in this paper, which lead to two distinct attention-based model architectures. The two architectures differ in where attention construction is applied, reflecting different design decisions between whether to construct attribute attention on individual input features (specialised) or using the concatenated feature tensor of all input feature vectors (shared). Comparing the interpretations generated by these two models helps us understand how these two attention mechanisms impact the model interpretability. We conduct experimental evaluation of the above models using a real-life dataset, and draw insights from comparative analysis between the proposed models and existing baselines for model accuracy and between the two different attention mechanisms for model interpretability.

Hence, the paper makes the following main contributions. 
\begin{itemize}
    \item It addresses the `black-box' problem in the context of predictive process analytics by developing interpretable models that use attention weights as an intrinsic model interpretation mechanism, so that they are capable of informing both \textit{what} and \textit{why} is a process prediction.
    \item It proposes an approach for building interpretable models that incorporate interpretability directly into the structure of a process predictive model. 
  
\end{itemize}

The paper is organised as follows. 
Section~\ref{sec:related} provides an overview of existing studies applying deep learning techniques to predictive process analytics along with a brief introduction to business process prediction and XAI.  
Section~\ref{sec:approach} details our approach for building interpretable models using different attention mechanisms. 
Section~\ref{sec:eval} presents the evaluation of the proposed models including discussion of the evaluation results. 
Finally, Section~\ref{sec:concl} summarises the contributions of our work and outline future work. 






\section{Background and Related Work}
\label{sec:related}

\subsection{\revision{Business Process Event Logs}} 
\label{subsec:bprocess}



A \textit{business process} is a set of coordinated activities performed to achieve certain business goals~\cite{Dumas_La_2013}. Organisations increasingly use information systems to support execution of their business processes. Such systems often record the execution of ongoing process events in \textit{event logs}, which can be used to analyse the behaviour and performance of business processes thus extracting insights for process improvement~\cite{van_der_aalst_process_2016}. 

Table~\ref{table:fragment} contains a fragment of a real-world event log \revision{from the dataset often referred to as \textit{BPIC~2012}~\cite{BPIC2012}. The complete event log captures the historical execution of process events for over $13,000$ loan applications in a Dutch financial institute over an about six-month period. It is used in the evaluation of our approach in Section~\ref{sec:eval}. Typically, such a loan application process starts with a customer submitting an application (\texttt{A\_SUBMITTED}) and ends with eventual conclusion of that application into an Approval (\texttt{A\_APPROVED}), Cancellation (\texttt{A\_CANCELLED}) or Rejection (\texttt{A\_DECLINED}).} 





\begin{table}[ht]
\centering
\resizebox{\columnwidth}{!}{
\begin{tabular}{lllll}
\hline
Case ID & Activity                                      & Timestamp               & Resource                  & \revision{Amount\_REQ} \\ \hline
173688  & A\_SUBMITTED                                  & 2011-10-01 08:38:44.546 & \revision{role\_}112      & 20000       \\
173688  & A\_PARTLYSUBMITTED                            & 2011-10-01 08:38:44.880 & \revision{role\_}112      & 20000       \\
173688  & A\_PREACCEPTED                                & 2011-10-01 08:39:37.906 & \revision{role\_}112      & 20000       \\
173694  & A\_SUBMITTED                                  & 2011-10-01 16:10:30.287 & \revision{role\_}112      & 7000        \\
173694  & W\_Filling in information for the application & 2011-10-01 19:35:59.637 & \revision{role\_}10912    & 7000        \\
173688  & A\_ACCEPTED                                   & 2011-10-01 19:42:43.308 & \revision{role\_}10862    & 20000       \\
173688  & O\_SELECTED                                   & 2011-10-01 19:45:09.243 & \revision{role\_}10862    & 20000       \\
173688  & A\_FINALIZED                                  & 2011-10-01 19:45:09.243 & \revision{role\_}10862    & 20000       \\
173688  & W\_Filling in information for the application & 2011-10-01 19:45:13.917 & NULL                      & 20000       \\
173694  & W\_Calling after sent offers                  & 2011-10-27 02:50:22.004 & \revision{role\_}11203    & 7000        \\
173694  & W\_Calling after sent offers                  & 2011-10-29 20:41:29.569 & \revision{role\_}11181    & 7000        \\
173694  & W\_Calling after sent offers                  & 2011-11-01 23:55:35.020 & \revision{role\_}11189    & 7000        \\
173694  & O\_ACCEPTED                                   & 2011-11-05 01:04:52.611 & \revision{role\_}10609    & 7000        \\
173694  & A\_APPROVED                                   & 2011-11-05 01:04:52.612 & \revision{role\_}10609    & 7000        \\
173694  & A\_REGISTERED                                 & 2011-11-05 01:04:52.612 & \revision{role\_}10609    & 7000        \\
173694  & A\_ACTIVATED                                  & 2011-11-05 01:04:52.612 & \revision{role\_}10609    & 7000        \\
173694  & W\_Assessing the application                  & 2011-11-05 01:05:01.532 & \revision{role\_}10609    & 7000        \\ \hline
\end{tabular}
}
\caption{A fragment of an event log of the loan application process recorded by \textit{BPIC~2012}~\cite{BPIC2012}}
\label{table:fragment}
\end{table}

In Table~\ref{table:fragment}, each row represents the execution of a process event, and each column represents an attribute of the event that carries relevant execution data. \revision{The events describe steps along the loan application process.} An instance of process execution, which comprises a sequence of events, is called a \textit{case}, and the sequence of events generated by a case forms a \textit{trace}. Each case is assigned a unique case identifier (e.g., to uniquely identify a loan application). For example, the sample log fragment in Table~\ref{table:fragment} records process execution data relevant to two loan applications \revision{among those captured by \textit{BPIC~2012}}. 

\revision{With regards to event attributes, each event has a unique identifier for the case to which the event belongs (\texttt{Case ID}), a name or label for the \textit{activity} of the event (\texttt{Activity}), and an execution time of the event (\texttt{Timestamp}). These are three mandatory attributes that an event needs to carry in a standard event log~\cite{xes_ieee_2016}. In the log fragment of \textit{BPIC~2012} shown in Table~\ref{table:fragment}, an event also has an attribute of a \textit{resource} name/identifier (\texttt{Resource}), indicating who performed the activity of the event. In addition, static attributes, such as loan amount (\texttt{Amount\_REQ})} which carry the same value for the events of the same case, may be included as well. 
\revision{For example, in Table~\ref{table:fragment} the event on the first row captures the execution of activity \texttt{A\_SUBMITTED} performed by resource \texttt{role\_112} at \texttt{08:38:44.546} on \texttt{2011-10-01} for loan application with reference number \texttt{173688} that requests a loan with amount of \texttt{20,000}.} 
Also, events in the same case (or trace) should be sequentially ordered by timestamps.

\begin{figure}[b!!!]
    \vspace*{-\baselineskip}
    \centering
    \includegraphics[width=0.7\textwidth]{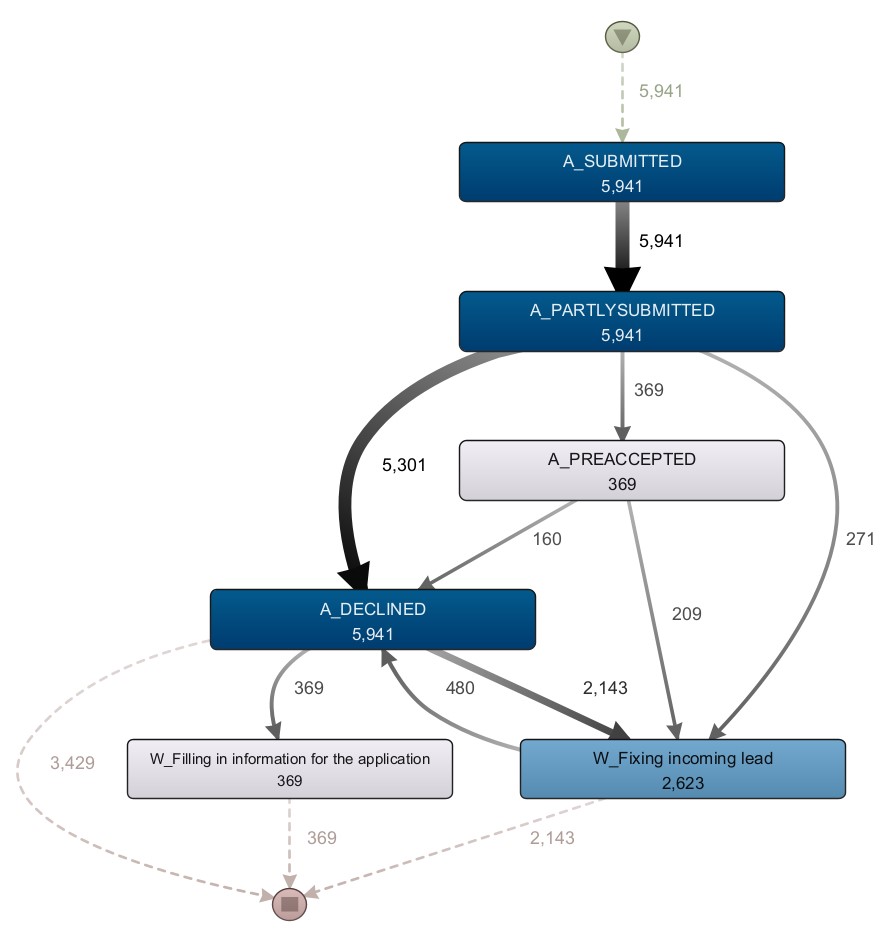}
    \caption{\revision{A process map depicting an abstract view of} the loan application process recorded by \textit{BPIC~2012}~\cite{BPIC2012}}
    \vspace*{-\baselineskip}
\label{fig:process_map_example}
\end{figure}

With an event log, we can extract correlations between activities and statistics of process execution captured by the log data. Process maps can be utilised, as a visual means, to present relevant process knowledge to human users. Typically, a process map is represented as a directed graph. For example, Figure~\ref{fig:process_map_example} shows a process map  illustrating an overview of the process execution traces captured by a portion of the aforementioned event log \textit{BPIC~2012}. The process map captures the execution traces of $5941$ loan application cases (which is a snapshot of about $45\%$ of the total cases in the complete event log) and is generated using a process mining tool Disco by Fluxicon\footnote{\url{https://fluxicon.com/disco}}. 
\revision{Note that the process map in Figure~\ref{fig:process_map_example} depicts an abstract view of the process for the purpose of illustrating core concepts of a business process and its execution event log. Only main activities and most frequent execution traces are shown in the figure. Process maps that depict the process execution of BPIC2012 at a much detailed level are included for discussion in Section~\ref{sec:eval}}.

In Figure~\ref{fig:process_map_example}, the triangular node inside a circle at the top of the process map symbolises the start of each case, and the square node inside a circle at the bottom of the map indicates the end of each case. Each rectangular node represents the execution of an activity and is labelled using the activity name (e.g., \texttt{A\_SUBMITTED}) and annotated with the execution frequency of the activity in terms of the number of cases involved (e.g., \texttt{A\_SUBMITTED} has occurred in all $5941$ cases). The rectangular nodes in Figure~\ref{fig:process_map_example} are also colour coded to visualise the level of activity execution frequency, e.g., dark blue indicates high frequency and light grey indicates low frequency along the corresponding colour spectrum. A directed arc connecting two nodes exhibits the execution path between the two activities and is annotated with the execution frequency of the path, where the thickness of the arc is used to visualise the frequency level. Also, a solid arc indicates a direct-follow execution order along the path, e.g., after the execution of \texttt{A\_SUBMITTED} the subsequent step is the execution of \texttt{A\_PARTLYSUBMITTED} (in all $5941$ cases).





\subsection{\revision{Business Process Prediction}} 
\label{subsec:bprediction}

Business process prediction focuses on analysing historical data in process event logs to predict future observations of a business process. These prediction tasks mainly involve the next-event forecasting~\cite{Evermann2017}, predicting the outcome of an ongoing case~\cite{teinemaa2019outcome}, and predicting remaining time for a case till its potential completion~\cite{verenich2019survey}. In the last decade, a variety of techniques have been used to conduct these predictions, and more recently, deep learning techniques such as LSTM have been adopted for many business process prediction tasks. 

Predicting the next event is a typical process prediction problem, and an important yet challenging topic in predictive process analytics. By applying next event prediction iteratively and progressively, it is possible to obtain a sequence of future events --- prediction of remaining sequence. This will ultimately lead to process completion resulting in process outcome prediction. The accuracy and reliability of the latter two predictions depend on the quality of next activity prediction. Therefore, our paper is mainly aimed at the next-event prediction task.




\subsection{Explainable AI} 

Explainable AI (XAI) techniques aim to help the humans to understand the underlying logic or criteria by which a machine learning model makes a decision. By nature, most of the machine learning models act as black-boxes, and despite the performance of the model, the understanding of model logic helps to improve the trust upon the model. Model explainability is expected to fulfil purposes of system verification, system improvement, and complying with legislation to provide such information when required~\cite{Samek2017}.

The \textit{post-hoc methods} of explaining a machine learning model (black-box) achieve their objective by computing a surrogate model that corresponds to a function that approximates the behaviour of the black-box. For example, LIME~\cite{Ribeiro2016} and SHAP~\cite{Lundberg2017} are well established post-hoc methods, that use the model inputs and outputs as the main material for constructing model explanations. LIME achieves interpretability by establishing a linear relationship between the model inputs and outputs using permutations of a single prediction. SHAP uses the Shapely values to explain the models which is a fundamental in game theory. Use of more interpretable surrogate models such as decision trees or regression models to approximate a black-box is also widely used, to achieve both the global (overall model logic) and local (one instance) level explanations~\cite{Giri2020,Boz2002}. The main challenge of applying these methods is, ensuring that these explanations are true to the model (explanation fidelity).

To overcome the challenge of explanation fidelity, the internal properties to explain a black-box can be used (intrinsic explanations). For whiter-boxes such as regression algorithms or decision trees these can be easily obtained by extracting the model coefficients or decision rules. However, for black-boxes such as deep learning models and complex ensemble machine learning algorithms, it is not as straight forward. Some of the existing approaches available in deep learning space include extracting regression activation maps (convolutional Neural Network specific)~\cite{Wol2020,Iad2021}, Partial dependence plots~\cite{Mehdiyev2020}, Layer wise relevance propagation~\cite{Samek2017,Wein2020}, and use of Attention weights (Recurrent Neural Network specific)~\cite{Xue2019}.
Improvising upon these techniques, the deep learning architectures can be further modified to facilitate the explainability~\cite{Retain_Choi2016,Retainvis_Kwon2019}, whilst not compromising on the model performance.

The evaluation of the effectiveness of a model explanation still remains an open question~\cite{Velmurugan2021}. Whilst there are no solid established measures of the effectiveness (that can be compared to metrics like model accuracy, recall that are used to measure the model performance), the primary indicators of a good model explanation are its fidelity (faithfulness to the actual model) and interpretability (if it makes sense for the humans)~\cite{XAIQuality_Zhou2021}.

\subsection{Deep Learning-based Predictive Process Analytics}  

Given the complex nature of the event logs, deep learning-based architectures offer a better information preservation compared to traditional machine learning models in predictive process analytics. In traditional machine learning approaches, due to the requirement of low complexity in feature vectors, it is required to encode the event log prefixes (partial process traces), in such a way that it might lose some of the information in the process. Amongst the deep learning-based architectures, RNN-based architectures have been a more popular choice, as they are designed to handle sequential inputs. Evermann et al.~\cite{Evermann2017,Evermann2017_2} and Tax et al.~\cite{Tax2017} introduced LSTM predictive models into business process predictions firstly. A more recent study~\cite{Camargo2019} has introduced three generalisable alternative LSTM-based architectures for next event prediction, which has given a significant inspiration to the work presented in this paper. 

Despite RNN-based architectures being the most popular option for business process prediction,many other deep learning techniques are explored in literature. Convolutional Neural Networks (CNNs)~\cite{Park2020, di_mauro_activity_2019, pasquadibisceglie_using_2019} were used in process predictions by encoding sequential event data in to spatial image-like data for process prediction. The modernistic research by Bukhsh et al.~\cite{bukhsh2021processtransformer} introduces a novel transformer framework into predictive process analytics, which replaces the RNN cells with attention mechanisms in recurrent neural network.

Most of the existing work in \textit{explainable} deep learning-based predictive process analytics approaches use post-hoc methods such as LIME and SHAP to explain the model's prediction~\cite{Sindhgatta2020a,Mehdiyev2020},whilst some of the recent approaches focus on intrinsic interpretable deep learning architectures. These include the approaches that use model attention~\cite{Sindhgatta2020a}, explicit process model and Gated graph neural network based approaches~\cite{Harl2020}, and use of partial dependence plots~\cite{Mehdiyev2021} and Layer wise relevance propagation~\cite{Samek2017}.


\paragraph{\bf Research Gap}

XAI is an area which offers significant opportunities in predictive process analytics and process mining as a whole, by assisting diagnose, fine-tune and improve predictive models, and using the insights to help improve the underlying process. 
However, in the existing literature, we find two key weaknesses. Not many interpretations are extracted in a way that will be meaningful in the business context. And there is little effort on understanding model explanations in relation to the process knowledge. 
These two weaknesses hinder the potential of existing XAI techniques as it is, to be used in a way that is beneficial to the domain. This is the main research gap we expect to address in our research.

\section{Approach}
\label{sec:approach}

\begin{figure}[h!]
    \centering
    \includegraphics[width=\textwidth]{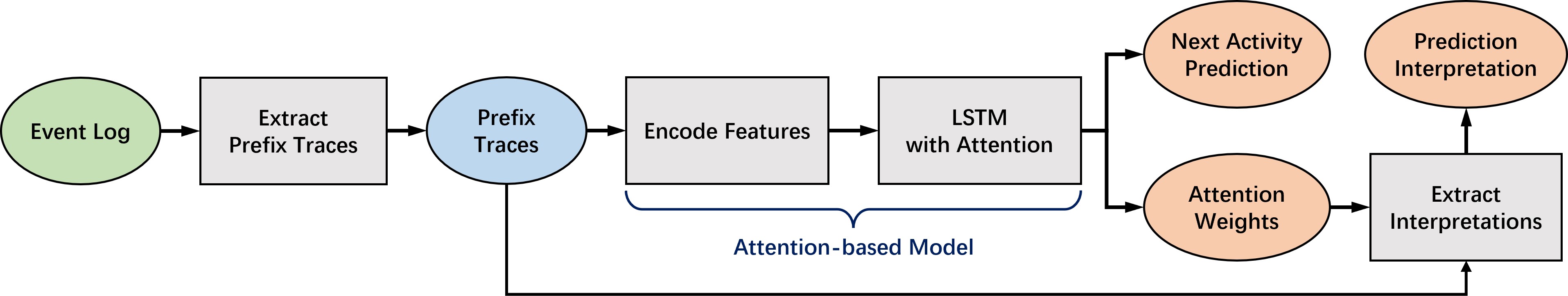}
    \vspace*{-.5\baselineskip}
    \caption{Overview of the approach}
\label{fig:approach_abs}
\end{figure}

This section presents our approach for constructing interpretable models for business process prediction. Figure~\ref{fig:approach_abs} provides an abstract illustration of the approach. 
Taking an \textit{event log} as the input, the data preparation mainly involves extracting \textit{prefix traces} from the event log (see Section~\ref{subsec:features}). The set of prefix traces are firstly converted into feature vectors and then fed into LSTM networks built with attentions. We consider both feature construction technique and LSTM network with attention as defining elements of the architecture of an \textit{attention-based model}, and propose two different attention mechanisms in the design of such a model (see Section~\ref{subsec:model}). Given a specific prefix trace to an attention-based model, the model is expected to predict the activity to occur in the subsequent step of the prefix trace, i.e., \textit{next activity prediction}, and also compute \textit{attention weights} when making the prediction. Finally, the model's attention weights and its input prefix trace are used to extract \textit{interpretations of model's prediction}, e.g., what features have made an impact and to what extent on the model's prediction (see Section~\ref{subsec:interpretations}).


\subsection{Event Logs and Prefix Traces}
\label{subsec:features}

An \textit{event log} records execution trails (a.k.a. instances) of a business process and provides the key input data for making process predictions. 
An instance of process execution is often referred to as a \textit{case}, and 
based on the notion of a \textit{trace} (which comprises a sequence of events of a case, see Section~\ref{subsec:bprocess}), a \textit{prefix trace} of length~$l$ contains the first~$l$ events of a trace. As such, multiple prefix traces of different lengths of a case capture the case execution \textit{progressively}, and are an important input to train a deep learning model for process prediction. Below, we formally define these key concepts and notions. 

\begin{myDef}[Event and attribute~\cite{Suriadi2015}]
\label{def:event}
\upshape 
Let $\SET{C}$ be the set of case identifiers, $\SET{A}$ the set of activity names, $\SET{R}$ the set of resource identifiers, and $\SET{T}$ the set of timestamps. $\SET{E}$ is the set of \textit{events}, and each event has the above \textit{attributes}\footnote{An event can have more attributes but these are not considered in this paper.}. For any $e\in\SET{E}$: $c_e\in\SET{C}$ is the case identifier of $e$, $a_e\in\SET{A}$ is the activity name of~$e$, $r_e\in\SET{R}$ is the resource identifier of~$e$, and $t_e\in\SET{T}$ is the timestamp of~$e$. If an attribute is missing, a $\bot$ value is returned, e.g., $r_e=\bot$ means that no resource is associated with event~$e$.
\end{myDef}

\begin{myDef}[Event log~\cite{Suriadi2015}]
\label{def:eventlog}
\upshape 
An \textit{event log} $\SET{L}\subseteq\SET{E}$ is a set of events. 
\end{myDef}

\begin{myDef}[Trace~\cite{van_der_aalst_process_2016}]
\label{def:trace}
\upshape 
A \textit{trace} $\sigma\subseteq\SET{L}$ is a non-empty sequence of unique events in $\SET{L}$. Let $n=|\sigma|$ and $\sigma=[e_1,...,e_n]$ (where positive integers~$1,...,n$~can be referred to as event index numbers). 
For all $i,j\in\{1,...,n\} \colon c_{e_i}=c_{e_j}$ (i.e., all events in a trace refer to the same case). 
Let $c_\sigma$ denote the case identifier of the case captured by trace~$\sigma$. For all $e\in\sigma \colon c_\sigma=c_e$, and for all $e'\in\SET{L}\setminus\sigma \colon c_\sigma \neq c_e'$ (i.e., each trace in an event log is uniquely identified by the case captured by the trace).
For $1 \leq i < j \leq n$: $e_i \neq e_j$ (i.e., each event appears only once), and $t_{e_i} \leq t_{e_j}$ (i.e., the ordering of events in a trace should respect their timestamps)\footnote{Event index numbers take precedence over timestamps where two events occur concurrently.}. 
\end{myDef}



\begin{myDef}[Prefix trace~\cite{teinemaa2019outcome}]
\label{def:prefixtrace}
\upshape 
Given a trace $\sigma=[e_1,...,e_n]$ and an integer $1\leq l\leq n$, $\s{prefix}(\sigma,l)=[e_1,...,e_l]$ is a \textit{prefix trace} of~$\sigma$ of length~$l$ (i.e., it contains the first~$l$ events of~$\sigma$).
\end{myDef}

\subsection{Attention-based Models} 
\label{subsec:model}

\begin{figure}[h!!!]
    \centering
    \includegraphics[width=\textwidth]{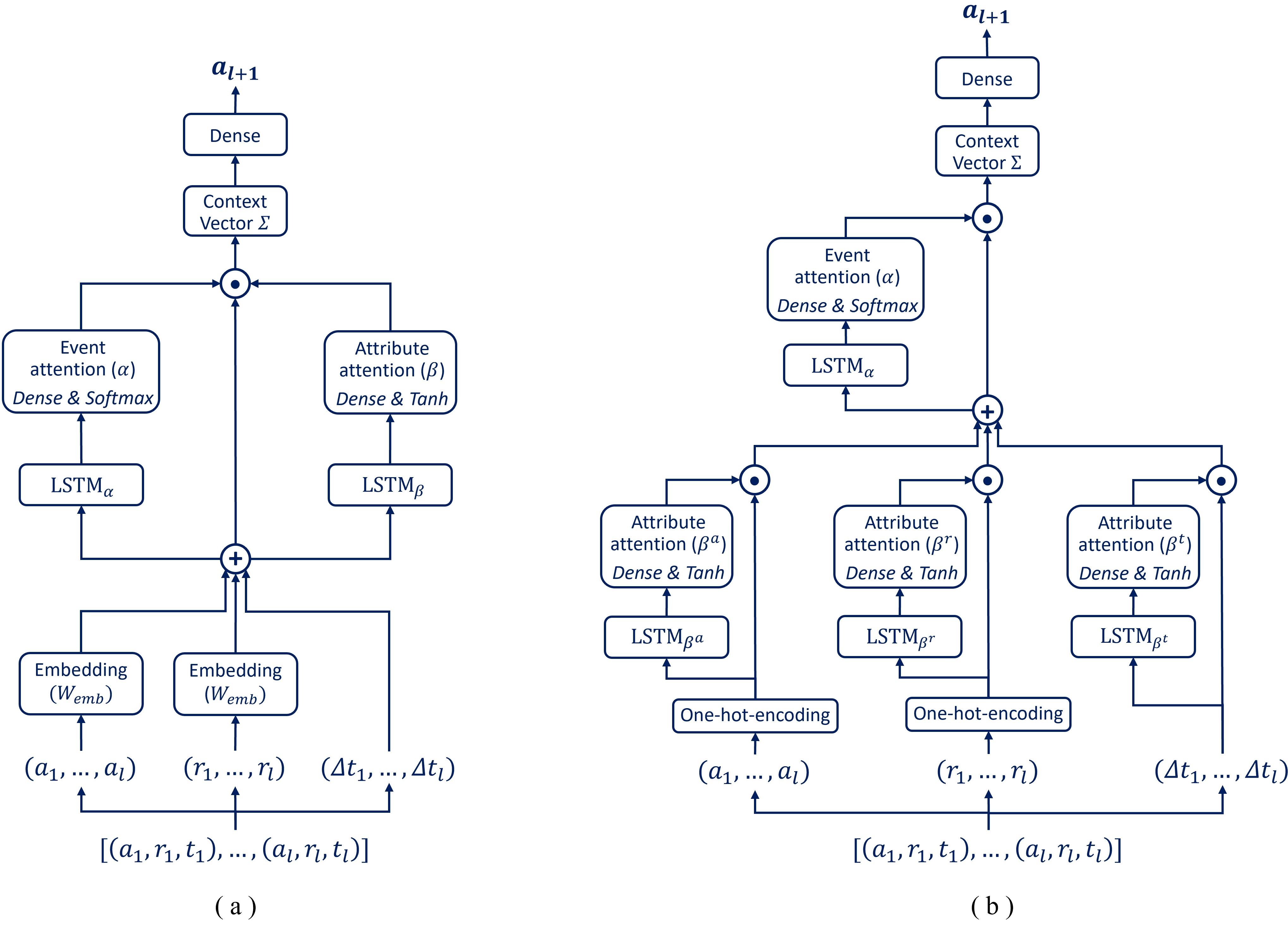}
    \vspace*{-1.5\baselineskip}
    \caption{Model architectures of (a) \textit{shared attention-based model} (based on~\cite{Sindhgatta2020a}) and (b) \textit{specialised attention-based model}} 
\label{fig:attention_models}
\end{figure}

%
We present the design of interpretable models for next activity prediction using two different attention mechanisms --- \textit{shared attention mechanism} and \textit{specialised attention mechanism}. These lead to two distinct attention-based model architectures as shown in Figure~\ref{fig:attention_models}. 
The shared attention-based model refers to the model architecture that has been proposed in our previous work~\cite{Sindhgatta2020a} and serves as the baseline model for comparison with the specialised attention-based model built upon the novel specialised attention mechanism proposed in this paper.


For simplicity, we describe the model input using the example of a single prefix trace. Based on the definitions in Section~\ref{subsec:features}, an event $e_i$ in a prefix trace can be represented as a tuple $(a_{e_i},r_{e_i},t_{e_i})$, or $(a_i,r_i,t_i)$ as a simplified notation. The time feature is specified to capture the time elapsed from the start event~$e_1$ to the current event~$e_i$ of a trace. We assume that each event has a timestamp associated with the completion of the event, and hence the time feature for event $e_i$ is computed as $\Delta t_i = t_i - t_1$. Given a prefix trace $[e_1,...,e_l]$, more specifically represented as $[(a_1,r_1,t_1),...,(a_l,r_l,t_l)]$, three feature vectors can be extracted as the input to a deep learning model, which are activity vector $(a_1,...,a_l)$, resource vector $(r_1,...,r_l)$, and a time interval vector $(\Delta t_1,...,\Delta t_l)$.

Both models take a prefix trace ($[(a_1,r_1,t_1),...,(a_l,r_l,t_l)]$) as the input
and predict the next activity ($a_{l+1}$) for the prefix trace. 
Bidirectional LSTM is applied as the deep learning network. Compared to a traditional LSTM which handles sequencial flow information in one direction, a bidirectional LSTM computes an additional set of hidden state vectors by considering reverse order of the input. The reverse order is useful to consider in the context of process execution, as the next activity often relies on the information about what has happened most recently (reverse order of activities). 

Both models support interpreting how the events and their attributes influenced the prediction by constructing the so-called \textit{event attention} ($\alpha$) and \textit{attribute attention} ($\beta$), respectively. 
\textit{Event attention} allows the model to focus on specific (event) indices in the prefix trace when predicting the next activity. Finding the indices in a prefix that contributes to a prediction can be derived based on $\alpha$ value --- the higher the value, the more influential the event at that index was in predicting the next activity. In our design, we apply \textit{softmax} as the activation function to compute event attention. 
In addition to identifying how events in a prefix trace influenced the prediction, it would also be useful to reason which of the (event) attributes influenced the prediction (e.g., activities, resources, or elapsed execution time). This has led to \textit{attribute attention} which can be derived based on $\beta$ which is a two dimensional vector. In our design, we apply \textit{tanh} as the activation function to compute attribute attention. We use \textit{tanh} as the activation function for attribute attention so that, for each feature value, it is clear which values would affect the decision negatively or positively along with the magnitude of the attention.  

The two model architectures differ in where attention weight construction is applied.

\begin{enumerate}[label=\roman*)]
    \item In the \textit{shared attention-based model} (Figure~\ref{fig:attention_models}(a)), both attentions are constructed using the concatenated embedded feature tensor which contains all three input feature vectors. Since the attention is shared amongst all three features, it is called as such. This model architecture is proposed based on our previous work~\cite{Sindhgatta2020a}. 
    \item In \textit{specialised attention-based model} (Figure~\ref{fig:attention_models}(b)): Three specialised attribute attention vectors are constructed on individual feature vector level. A separate event attention vector is constructed using both attention vectors and feature vectors as inputs, to obtain the event level influence to the final prediction.
\end{enumerate}

Below, we define the construction of each model with the aid of mathematical notations where relevant.

\subsubsection{Shared Attention-based Model}

Figure~\ref{fig:attention_models}(a) depicts the model that predicts the next activity of an input prefix trace using \textit{shared attention mechanism}. 
For a give prefix trace, the input sequences of activities $(a_1,...,a_l)$, resource $(r_1,...,r_l)$ and time elapsed $(\Delta t_1,...,\Delta t_l)$, are initially converted into embedded vectors denoted as follows.

\begin{center}
    $\vect{v}^\s{emb}_a = \mathbf{W}^\s{emb}_a~(a_1,...,a_l)$ 

    $\vect{v}^\s{emb}_r = \mathbf{W}^\s{emb}_r~(r_1,...,r_l) $
    
    $\vect{v}_t = (\Delta t_1,...,\Delta t_l)$
\end{center}

The three vectors are then concatenated into a two dimensional embedded feature vector as follows.
\begin{center}
    $\vect{v}^{emb}_{a,r,t} = \vect{v}^\s{emb}_a \cup \vect{v}^\s{emb}_r \cup \vect{v}_t$,~where $\vect{v}^{emb}_{a,r,t} \in \mathbb{R}^{l\times(|\SET{A}|+|\SET{R}|+1)}$
\end{center}

The same feature vector is then passed through two independent bidirectional LSTMs ($LSTM_\alpha,LSTM_\beta$). Compared to a traditional LSTM which processes sequence information in one direction or from the first input in the sequence to the last input, a bidirectional LSTM computes an additional set of hidden state vectors by considering reverse order of the input as well. This is important in the context of process, as the most recently occurred events could have a higher bearing towards the prediction.

In order to compute \textbf{event attention $\alpha$}, $\mathit{LSTM}_\alpha$ generates the hidden state vector~$h \in \mathbb{R}^{l \times(|\SET{A}|+|\SET{R}|+1)}$, taking the embedded feature vector $\vect{v}^{emb}_{a,r,t}$ as the input. That is,   
\begin{center}
    $h \leftarrow \mathit{LSTM}_\alpha( \vect{v}^{emb}_{a,r,t})$
\end{center}
%
Then, event attention $\alpha\in\mathbb{R}^l$ is computed via an immediate \textit{softmax} dense layer as follows. 
\begin{center}
    $\alpha_i = \mathit{softmax}( W^T_\alpha h_{ij} + b_\alpha)$ 
\end{center}
where $i \in \{1,...,l\}$, $j \in \{1,...,|\SET{A}|+|\SET{R}|+1\}$, and $W_\alpha$ and $b_\alpha$ are the trained weight matrix and bias of the \textit{softmax} dense layer, respectively.

In order to compute \textbf{attribute attention $\beta$}, $\mathit{LSTM}_\beta$ generates the hidden state vector $g \in \mathbb{R}^{l \times(|\SET{A}|+|\SET{R}|+1)}$, taking the embedded feature vector $\vect{v}^{emb}_{a,r,t}$ as the input. That is, 
\begin{center}
    $g \leftarrow \mathit{LSTM}_\beta( \vect{v}^{emb}_{a,r,t})$ 
\end{center}
%
Then, attribute attention $\beta \in \mathbb{R}^{l\times(|\SET{A}|+|\SET{R}|+1)}$ is computed via an immediate~\textit{tanh} dense layer as follows. 
\begin{center}
    $\beta_{ij} = \mathit{tanh}( W_\beta g_{ij} + b_\beta)$ 
\end{center}
where $i \in \{1,...,l\}$, $j \in \{1,...,|\SET{A}|+|\SET{R}|+1\}$, and $W_\beta$ and $b_\beta$ are the trained weight matrix and bias of the \textit{tanh} dense layer, respectively.

The attention values of both event attention~$\alpha$ and attribute attention~$\beta$ are then element-wise multiplied with the concatenated feature vector $\vect{v}^{emb}_{a,r,t}$ and summed across the entire input prefix trace (of length~$l$) to obtain a context vector $c$. That is,  
\begin{center}
    $c = \sum^l \alpha \odot\beta \odot \vect{v}^{emb}_{a,r,t}$ 
\end{center}
which represents the cumulative effect of event and attribute level influence. This context vector then goes through a dense layer to calculate the final prediction 
--- the next activity $a_{l+1}$ of the input prefix trace.

\subsubsection{Specialised Attention-based Model} 

Figure~\ref{fig:attention_models}(b) depicts the specialised attention-based model, where activity and resource feature vectors are each converted into one-hot-encoded two dimensional feature vectors, i.e.,\ 
\begin{center}
    $(a_1,...,a_l) \rightarrow \vect{v}^\s{ohe}_a \in \{0,1\}^{l\times|\SET{A}|}$

    $(r_1,...,r_l) \rightarrow \vect{v}^\s{ohe}_r \in \{0,1\}^{l\times|\SET{R}|}$
\end{center}
and time elapsed feature, which is a real numbered feature vector, is used as it is, i.e., $\vect{v}_t = (\Delta t_1,...,\Delta t_l)$. 

In order to compute \textbf{attribute attention $\beta$}, each feature vector is passed through the corresponding individual BiLSTMs --- $\mathit{LSTM}_{\beta_a}$, $\mathit{LSTM}_{\beta_r}$, and $\mathit{LSTM}_{\beta_t}$ --- to generate the hidden state vectors $g^a\in \mathbb{R}^{l\times|\SET{A}|}$, $g^r \in \mathbb{R}^{l\times|\SET{R}|}$, and $g^t \in \mathbb{R}^l$, respectively.
\begin{center}
    $g^a \leftarrow \mathit{LSTM}^a_\beta( \vect{v}^\s{ohe}_a)$ 
    
    $g^r \leftarrow \mathit{LSTM}^r_\beta( \vect{v}^\s{ohe}_r)$ 
    
    $g^t \leftarrow \mathit{LSTM}^t_\beta( \vect{v}_t)$ 
\end{center}
Attribute attention vectors for each feature category $\beta^a\in \mathbb{R}^{l\times|\SET{A}|}$, $\beta^r \in \mathbb{R}^{l\times|\SET{R}|}$, and $\beta^t \in \mathbb{R}^l$ are then computed correspondingly via three individual immediate \textit{tanh} dense layers as follows.
\begin{center}
   $\beta^a_{ij} = \mathit{tanh}(W^a_\beta g^a_{ij} + b^a_\beta)$ 
\end{center}
where $i \in \{1,...,l\}$, $j \in \{1,...,|\SET{A}|\}$, and $W^a_\beta$ and $b^a_\beta$ are the trained weight matrix and bias of the \textit{tanh} dense layer used to compute $\beta^a$, respectively.
\begin{center}
   $\beta^r_{ij} = \mathit{tanh}(W^r_\beta g^r_{ij} + b^r_\beta)$ 
\end{center}
where $i \in \{1,...,l\}$, $j \in \{1,...,|\SET{R}|\}$, and $W^r_\beta$ and $b^r_\beta$ are the trained weight matrix and bias of the \textit{tanh} dense layer used to compute $\beta^r$, respectively.
\begin{center}
   $\beta^t_{i} = \mathit{tanh}(W^t_\beta g^t_{i} + b^t_\beta)$ 
\end{center}
where $i \in \{1,...,l\}$, and $W^t_\beta$ and $b^t_\beta$ are the trained weight matrix and bias of the \textit{tanh} dense layer used to compute $\beta^t$, respectively.

By multiplying the attention vectors with the relevant feature vectors, we then arrive at an intermediate vector which represents the `influence' of each feature at feature specific level.
\begin{center}
    $\vect{v}^\s{inf}_a = \vect{v}^\s{ohe}_a\odot\beta^a$

    $\vect{v}^\s{inf}_r =\vect{v}^\s{ohe}_r\odot\beta^r$
    
    $\vect{v}^\s{inf}_t = \vect{v}_t\odot\beta^t$
\end{center}
The concatenation of the three feature level influence vectors is then passed through a single BiLSTM layer to obtain the event attention $\alpha$ by considering the influence of all three features simultaneously as follows. 
\begin{center}
    $\vect{v}^\s{inf}_{a,r,t} = \vect{v}^\s{inf}_a \cup \vect{v}^\s{inf}_r \cup \vect{v}^\s{inf}_t$,~where $\vect{v}^\s{inf}_{a,r,t} \in \mathbb{R}^{l\times(|\SET{A}|+|\SET{R}|+1)}$ 
\end{center}

Next, we compute \textbf{event attention $\alpha$}. $\s{LSTM}_\alpha$ generates the hidden state vector $h \in \mathbb{R}^{l \times (|\SET{A}|+|\SET{R}|+1})$, taking the concatenated feature level influence vector $\vect{v}^\s{inf}_{a,r,t}$ as the input. That is, 
\begin{center}
    $h \leftarrow \s{LSTM}_\alpha( \vect{v}^\s{inf}_{a,r,t})$ 
\end{center}
Event attention $\alpha\in\mathbb{R}^l$ is then computed via an immediate \textit{softmax} dense layer as follows. 
\begin{center}
    $\alpha_i = \s{softmax}(W^T_\alpha h_{ij} + b_\alpha)$ 
\end{center}
where $i \in \{1,...,l\}$, $j \in \{1,...,|\SET{A}|+|\SET{R}|+1\}$, and $W_\alpha$ and $b_\alpha$ are the trained weight matrix and bias of the \textit{softmax} dense layer, respectively.

Finally, the above event attention~$\alpha$ is element-wise multiplied with the concatenated feature influence vector~$\vect{v}^{inf}_{a,r,t}$, and summed over the entire prefix to obtain a context vector $c$, which represents the cumulative effect of event and attribute level influence. This context vector then goes through a dense layer to calculate the final prediction --- the next activity $a_{l+1}$ of the input prefix trace. 
\begin{center}
    $c = \sum^l \alpha \odot \vect{v}^{inf}_{a,r,t}$
\end{center}

\subsection{Extracting Interpretations}
\label{subsec:interpretations}


In our approach, we extract the feature contributions (the weight of each individual feature bears relative to the other features) using attention weights for interpretations. Such interpretations can be used to explain why a model makes a certain prediction, specifically what input features have impacted the model prediction and to what extent.

By design, in both models, we compute event attention~$\alpha$ and (event) attribute attention~$\beta$. As the names indicate, $\alpha$ gives the influence of each time step (i.e., event) of a prefix trace towards the final prediction, whilst a combination of $\alpha$ and $\beta$ gives the influence (i.e.,  feature weight) of each individual event attribute towards prediction.

To extract the feature weights, we first multiply attention vector~$\alpha$ with attention vector~$\beta$ element wise. This yields a combined attention-based feature weight matrix, built upon the attention weights. The weight extraction mechanism from this matrix is different between shared attention-based model and specialised attention-based model.

In shared attention-based model, to obtain the feature weights of a given event~$e_i$ we use the indexes that were given to the features of event attributes~$(a_i,r_i,t_i)$ when they were index encoded during embedding. Let $j^a$ denote the index of $a_i$ and $j^r$ the index of $r_i$, then the feature weights of event attributes $a_i$, $r_i$, and $t_i$ can be obtained respectively as follows. 

\begin{center}
    feature weight of $a_i$ = value of the ($j^a$)th element of $\alpha_i \odot \beta_i$ \\
    feature weight of $r_i$ = value of the ($|\SET{A}|+j^r$)th element of $\alpha_i \odot \beta_i$ \\
    feature weight of $t_i$ = value of the ($|\SET{A}|+|\SET{R}|+1$)th element of $\alpha_i \odot \beta_i$ \\
\end{center}

In specialised attention-based model, one-hot-encoding allows for orthogonal relationship between distinct feature values. Therefore, in the combined feature weight matrix, for a given event~$e_i$, 
only the relevant feature values carry a non-zero weight. Thus, the feature weights can be extracted straightforward as follows.

\begin{center}
    feature weight of $a_i =$ non-zero value of $\alpha_i \odot \beta^a_i$ \\ 
    feature weight of $r_i =$ non-zero value of $\alpha_i \odot \beta^r_i$ \\
    feature weight of $t_i =$ value of $\alpha_i \odot \beta^t_i$
\end{center}

Our approach supports extracting feature weights of events and their event attributes at all individual time steps (as explained above). In practice, it is often useful to focus on selected events and their attributes for interpretation and analysis. For example, one can identify $k$ most influential time steps (i.e., events) of an input prefix trace for which the model pays the highest attention to, by comparing the attention weights in the output of event attention ($\alpha$) layer. Once the $k$ most influential time steps are identified, the feature weights of event attributes ($\beta$ attention) pertaining to each corresponding time step can be extracted. This will be discussed in more detail in Section~\ref{subsec:interpretations-analysis}.   



\section{Evaluation}
\label{sec:eval}

\subsection{Experimental Design}

\paragraph{\bf Dataset}

The dataset used in the experiment contains event logs that belong to a loan application process of a Dutch Financial Institute. 
\revision{These event logs are relevant to the above process execution in different time frames, and were made publicly available for the annual business process challenge at the 8th and the 13th International Workshops of Business Process Intelligence in 2012 and 2017, respectively. They are often referred to as \textit{BPIC 2012}~\cite{BPIC2012} and \textit{BPIC 2017}~\cite{Dongen2017} event logs. Note that \textit{BPIC 2012} event log and the associated loan application process are introduced as an example in Section~\ref{subsec:bprocess} and hence some detailed description can be referred to in the section.
\textit{BPIC 2017} is an event log that records the execution of an evolved version of the loan application process in the same Dutch financial institute as that of \textit{BPIC 2012}. Both event logs contain process execution data} 
pertaining to three sub-processes. These are: Application sub-process (denoted BPIC~2012~$A$), Offer sub-process (BPIC~2012~$O$) and Workflow (automated) sub-process (BPIC~2012~$W$). 
Table~\ref{table:eventlog} shows the statistics for the data profiles of the event logs used in the experiment. 
%
%
 
\begin{table}[htp!]
\centering
\resizebox{\columnwidth}{!}{%
\begin{tabular}{lcccccccc} 
\hline
Event Log & \begin{tabular}[c]{@{}c@{}}Num.\\ cases\end{tabular} & \begin{tabular}[c]{@{}c@{}}Num.\\ activities\end{tabular} & \begin{tabular}[c]{@{}c@{}}Num.\\ event\end{tabular} & \begin{tabular}[c]{@{}c@{}}Avg.\\ case length\end{tabular} & \begin{tabular}[c]{@{}c@{}}Max.\\ case length\end{tabular} & \begin{tabular}[c]{@{}c@{}}Avg.\\ case duration\end{tabular} & \begin{tabular}[c]{@{}c@{}}Max.\\ case duration\end{tabular} & Variants \\ \hline
BPIC 2012 & 13087 & 36 & 262200 & 20.04 & 175 & 8.62 days & 137.22 days &
4366 \\
BPIC 2012 Complete & 13087 & 23 & 164506 & 12.57 & 96 & 8.61 days &
91.46 days & 4336 \\
BPIC 2012 W Complete & 9658 & 6 & 72413 & 7.5 & 74 & 11.4 days & 91.04 days
& 2263 \\
BPIC 2012 O & 5015 & 7 & 31244 & 6.23 & 30 & 17.18 days & 89.55 days & 168
\\
BPIC 2012 A & 13087 & 10 & 60849 & 4.65 & 8 & 8.08 days & 91.46 days & 17 
\\
\revision{BPIC 2017} & \revision{31509} & \revision{26} & \revision{1160405} & \revision{36.8} & \revision{177} & \revision{21.9 days} & \revision{286.04 days} & \revision{15484} 
\\\hline

\end{tabular}
}
\caption{Data profiles of event logs used in experiments.}
\label{table:eventlog}
\end{table}

An event log comprises of both dynamic and static Features. Whilst dynamic features change over the execution of the process, static features are case specific and remains constant throughout a case.  Table~\ref{table:features} depicts the features that we have considered in our experiments.

\begin{table}[!ht]
\centering
\resizebox{0.6\columnwidth}{!}{
\begin{tabular}{lll}
\hline
\multicolumn{2}{l}{Feature Category}                    & BPIC 2012          \\ \hline
\multicolumn{2}{l}{Single Process Trace Identification} & Case\_ID           \\
\multicolumn{2}{l}{Dynamic Features}                    &                    \\
                  & Activity                            & Concept\_Name       \\
                  & Role (Performer)                    & Resource           \\
                  & Timestamp                           & Complete\_Timestamp \\
                  & Lifecycle Transition                & Completed          \\
\multicolumn{2}{l}{Static Features}                     & None               \\ \hline
\end{tabular}
}
\caption{Event log features considered for the experiment.}
\label{table:features}
\end{table}

\paragraph{\bf Prefix Generation} 
To predict the next event(s) of a trace of ongoing process execution, we construct process execution prefix traces using the input event log data to train the model. We achieve this by firstly generating execution traces for each case (unique identified by Case ID). For each Case ID, we only consider the activities which is at `Completed' life-cycle transition for this purpose. Next, we organise all the dynamic features by `Timestamp' field in ascending order to generate an execution trace for a given Case ID. 
Then we generate prefixes at each activity, by considering the partial trace from the start of the trace to the given activity. Some of these prefixes may contain activity level self loops, i.e., the same activity being executed multiple times consecutively. However, at event level, when we take all the event attributes in consideration (activity, resource and timestamp), each of these occurrences represent a unique event in the process trace. Therefore, we consider each of those events as it is in the prefixes that are fed to the model.

\paragraph{\bf Data Pre-processing} 
Once the prefix traces are generated, the raw dataset needs to be prepared for the prediction task. We are interested in those features that represent the following attributes of a single event in a prefix trace. The `activity label' ($a$- categorical), `role' ($r$ - categorical) that preformed the event and `event complete timestamp' ($t$ - date time). These features were converted into numerical features by using index encoding for the categorical features, and obtaining the time elapsed $\Delta t$ (i.e., the time interval) between the event complete timestamp of the first event of the prefix trace and the event complete timestamp of the given event. Each prefix of length~$l$ is then converted into a two dimensional feature matrix identified by the prefix trace ID. The dimensions of the feature matrix are the event number ($i$) and the three features that identifies an event ($a_i, r_i, \Delta t_i$). The feature matrix can be represented as 
$([a_1, …, a_l], [r_1, …, r_l], [\Delta t_1, …, \Delta t_l])$. The entire pre-processed dataset (of prefixes) can be represented as a 3-dimensional feature tensor, where the third dimension is the prefix ID.

\paragraph{\bf Model Parameters}

The key building blocks of the shared attention-based model and specialised attention-model are depicted in Table~\ref{table:shared_attn_model} and~\ref{table:specialized_attn_model}, respectively.

\begin{table}[!ht]
\centering
\resizebox{\columnwidth}{!}{
\begin{tabular}{lllll}
\hline
\textbf{Layer}       & \textbf{Layer Type} & \textbf{Layer Parameters} & \textbf{Activation Function} & \textbf{Connected to}                                          \\ \hline
\textit{\textbf{$W_{emb}^a$}}                & Embedding           &                           &                              & \begin{tabular}[c]{@{}l@{}}LSTM $\alpha$\\ LSTM $\beta$\end{tabular} \\
\textit{\textbf{$W_{emb}^r$}}                & Embedding           &                           &                              & \begin{tabular}[c]{@{}l@{}}LSTM $\alpha$\\ LSTM $\beta$\end{tabular} \\
\textit{\textbf{LSTM $\alpha$}}               & Bidirectional LSTM  & LSTM size = 50            &                              & Event attention $\alpha$                                          \\
\textit{\textbf{LSTM $\beta$}}                & Bidirectional LSTM  & LSTM size = 50            &                              & Attribute attention $\beta$                                       \\
\textit{\textbf{Event Attention $\alpha$}}    & Dense               &                           & Softmax                      & Context                                                        \\
\textit{\textbf{Attribute Attention $\beta$}} & Dense               &                           & Tanh                         & Context                                                        \\
\textit{\textbf{Context}}                  & Multiply            &                           &                              & Dense                                                          \\
\textit{\textbf{Dense}}                    & Dense               &                           & Softmax                      & Final Output                                                   \\ \hline
\end{tabular}}
\caption{\revision{Elements of shared attention-based model}}
\label{table:shared_attn_model}
\end{table}

\begin{table}[!ht]
\centering
\resizebox{\columnwidth}{!}{
\begin{tabular}{lllll}
\hline
\textbf{Layer}                                & \textbf{Layer Type} & \textbf{Layer Parameters} & \textbf{Activation Function} & \textbf{Connected to}                                         \\ \hline
\textit{\textbf{$W_{ohe}^a$}}                   & One-hot-encoding    &                           &                              & LSTM $\beta^a$                                                  \\
\textit{\textbf{$W_{ohe}^r$}}                   & One-hot-encoding    &                           &                              & LSTM $\beta^r$                                                  \\
\textbf{LSTM $\beta^a$}                         & Bidirectional LSTM  & LSTM size = 50            &                              & Attribute attention $\beta^a$                                   \\
\textbf{LSTM $\beta^r$}                         & Bidirectional LSTM  & LSTM size = 50            &                              & Attribute attention $\beta^r$                                   \\
\textbf{LSTM $\beta^t$}                         & Bidirectional LSTM  & LSTM size = 50            &                              & Attribute attention $\beta^t$                                   \\
\textit{\textbf{LSTM $\alpha$}}                  & Bidirectional LSTM  & LSTM size = 50            &                              & Event attention $\alpha$                                         \\
\textit{\textbf{Attribute attention $\beta^a$}} & Dense               &                           & Tanh                         & \begin{tabular}[c]{@{}l@{}}LSTM $\alpha$ \\ Context \end{tabular} \\
\textit{\textbf{Attribute attention $\beta^r$}} & Dense               &                           & Tanh                         & \begin{tabular}[c]{@{}l@{}}LSTM $\alpha$ \\ Context \end{tabular} \\
\textit{\textbf{Attribute attention $\beta^t$}} & Dense               &                           & Tanh                         & \begin{tabular}[c]{@{}l@{}}LSTM $\alpha$ \\ Context \end{tabular} \\
\textit{\textbf{Event attention $\alpha$ }}       & Dense               &                           & Softmax                      & Context                                                       \\
\textit{\textbf{Context}}                     & Multiply            &                           &                              & Dense                                                         \\
\textit{\textbf{Dense}}                       & Dense               &                           & Softmax                      & Final Output                                                  \\ \hline
\end{tabular}}
\caption{\revision{Elements of specialised attention-based model}}
\label{table:specialized_attn_model}
\end{table}

\revision{In terms of hyperparameters for} the~\textit{shared attention-based model}~\cite{Sindhgatta2020a}, we maintain the original hyperparameters proposed by the authors. With the objective of mimicking the~\textit{shared attention-based model} except for the explicit architectural differences, in~\textit{specialised attention-based model}, we closely adapt the hyperparameters from the former model. In this paper, we do not explore hyperparameter optimisation~\cite{Kaselimi2019} to further improve the models, and this can be investigated in our future work. 

\paragraph{\bf Evaluation Metrics of Model Performance and Benchmarks} Model performance is measured by Accuracy, Precision, Recall and F1-score. \revision{Given the confusion matrix (see Table~\ref{table:confusion_matrix}) which categorises the prediction results based on the actual data class (ground truth), the computation of model performance metrics is as per Table~\ref{table:eval_metrics}.} 
We also compare our model performance against the state-of-the-art benchmark studies for next activity prediction in business processes~\cite{RamaManeiro2020} using Accuracy as the measure. 

\color{blue}

\begin{table}[htp!]
\centering
\resizebox{0.6\columnwidth}{!}{
\begin{tabular}{|l|cc|}
\hline
Data Class & \multicolumn{1}{c|}{Classified as Positive} & Classified as Negative \\ \hline
Positive   & True Positive ($t_p$)                             & False Negative ($f_n$)        \\ \cline{1-1}
Negative   & False Positive ($f_p$)                               & True Negative ($t_n$)     \\ \hline
\end{tabular}}
\caption{\revision{Confusion Matrix~\cite{SOKOLOVA_2009}}}
\label{table:confusion_matrix}
\end{table}

\begin{table}[htp!]
\centering
\resizebox{\columnwidth}{!}{
\begin{tabular}{l|cl}
\hline
                            & Formula & Purpose of the Metric                                                          \\ \hline
\textit{\textbf{Accuracy}}  & $\frac{t_p + t_n}{t_p + t_n + f_p + f_n}$   & Overall effectiveness of the classifier                                        \\
\textit{\textbf{Precision}} & $\frac{t_p}{t_p + f_p}$   & Class agreement of the positive labels being predicted                         \\
\textit{\textbf{Recall}}    & $\frac{t_p}{t_p + f_n}$   & How well does the classifier predict positive labels                           \\
\textit{\textbf{F1-Score}}  & $\frac{t_p}{t_p + \frac{1}{2} (f_p + f_n)}$   & Relations between the data's positive labels and those given by the classifier \\ \hline
\end{tabular}}
\caption{\revision{Evaluation Metrics~\cite{SOKOLOVA_2009}}}
\label{table:eval_metrics}
\end{table}

\color{black}

\paragraph{\bf Testing and Validation} For the purpose of testing the model performance, the dataset was randomised by the prefix trace ID and divided into two parts initially, training (contained $70\%$ of the prefixes) and testing ($30\%$ of the prefixes). To validate the consistency of the model performance, multiple experiments were run, randomizing the data set, and retraining the model and testing for the model performance.

\paragraph{\bf Evaluation of Model Interpretations} 
To measure the quality of a model interpretation, there are no established performance measures. The quality of a model interpretation can however be assessed by considering two properties of an interpretation. How true it is to the decision making process of the underlying model (fidelity) and how much does the interpretation make sense to the humans; lay people or domain experts (interpretability)~\cite{XAIQuality_Zhou2021}. Since we extract the interpretations from the intrinsic properties of the model itself, it will be true to the model (fidelitious or faithful). 
We evaluate the interpretability property of the model interpretations by comparing those interpretations to the insights gained from the domain knowledge. For this purpose we use domain expert reports related to the process captured by our evaluation Dataset~\cite{Bautista2013} and process mining tools such as ProM\footnote{\url{http://www.promtools.org/doku.php?id=prom611}} and Disco by Fluxicon (see Section~\ref{subsec:bprocess}).

For better comparison against the domain knowledge, we extract interpretations for the predictions that are made at key decision points of the process. A key decision point is defined by a crucial activity, after which the process can get directed to multiple directions (thus can have multiple next event targets). We select two such decision points for evaluation of interpretations, one early in the process, and another at the middle of the process. We analyse the behaviour of model interpretations at each of these decision points and what we can learn from the extracted interpretations.

\paragraph{\bf Implementation Details} The experiments were performed on a server with Windows 10 Operation System and its hardware contained 3.8 GHz AMD Ryzen 3900X CPU having 64 GB RAM and one single NVIDIA RTX A4000 GPU with 16 GB Memory. Prefix generation was performed using Structured Query Language (SQL) on the platform of Microsoft SQL server management studio. Both shared and specialised attention-based models are implemented with TensorFlow 2.5 library in Python. The constructed models are trained by an ADAM optimiser with a learning rate of 0.001 for all event logs in the experiments.For the purpose of reproducibility, the source code for experiments can be downloaded from \href{https://git.io/JE9fT}{\textbf{\textit{https://git.io/JE9fT}}}.

\subsection{Model Performance}

Table~\ref{table:metrics} presents the model performance of the shared attention-based model and specialised attention-based model for predicting next activity using \revision{five} event logs sourced from the \textit{BPIC~2012} \revision{and \textit{BPIC~2017}} dataset. The model performance measures are accuracy, precision, recall and f1-score. For all \revision{five} event logs, the performance of the specialised attention-based model is either the same as or slightly lower than that of the shared attention-based model. In the latter case, the difference is not statistically significant (see Appendix~A.1), which means that the performance of the two models can be considered similar to each other. 

\begin{table}[htp!]
\centering
\resizebox{\columnwidth}{!}{
\begin{tabular}{lcccccccc}
\hline
                    & \multicolumn{2}{c}{Accuracy} & \multicolumn{2}{c}{Precision} & \multicolumn{2}{c}{Recall} & \multicolumn{2}{c}{F1-score} \\ \cline{2-9} 
                    & Shared      & Specialised      & Shared       & Specialised      & Shared     & Specialised     & Shared      & Specialised      \\ \hline
BPIC 2012 Complete   & 0.79          & 0.78         & 0.77           & 0.75         & 0.79         & 0.78        & 0.74          & 0.73         \\
BPIC 2012 A          & 0.75          & 0.75         & 0.67           & 0.74         & 0.75         & 0.75        & 0.69          & 0.70         \\
BPIC 2012 O          & 0.82          & 0.82         & 0.82           & 0.81         & 0.82         & 0.82        & 0.80          & 0.80         \\
BPIC 2012 W Complete & 0.84          & 0.84         & 0.85           & 0.85         & 0.84         & 0.84        & 0.84          & 0.83         \\ 
\revision{BPIC 2017}            & \revision{0.83}         &  \revision{0.82}         & \revision{0.83}           & \revision{0.82}         & \revision{0.83}          & \revision{0.82}         &   \revision{0.82}     & \revision{0.81}            \\
\hline
\end{tabular}
}
\caption{The performance of the \textit{shared attention-based model} and \textit{specialised attention-based model}.}
\label{table:metrics}
\end{table}

In addition, we also compared the accuracy of our models against those from recent benchmark study~\cite{RamaManeiro2020}.\footnote{\revision{Note that only BPIC 2012 dataset was used in the benchmark studies~\cite{RamaManeiro2020}.}} 
Table~\ref{table:benchmark} reports the performance comparison with related research efforts on next-activity prediction. It can be observed that the shared and specialised attention-based mechanisms both achieved a reasonably good level of model accuracy. Especially they surpass other approaches at predicting next event in \textit{BPIC 2012 W Complete}, 
reaching to 84.39\% in shared attention-based mechanisms and 83.75\% in specialised attention-based mechanisms, respectively. 
In contrast to the approaches in that benchmark study, which only select activities and roles as input attributes for predictive models, both our models take the elapsed time as an additional input (time) attribute into consideration. Eventually, our models can learn more insights from the time features and increase model accuracy. Despite our approach not outperforming in the dataset \textit{BPIC 2012 Complete}, \textit{BPIC 2012 A} and \textit{BPIC 2012 W}, the model accuracy is at a comparable level of the benchmark results. It is worth pointing out that our models are not aimed at accurate prediction only, instead we would like to achieve a great balance between accuracy and interpretability.

\begin{table}[htp!]
\centering
\resizebox{\columnwidth}{!}{
\begin{tabular}{lcccc}
\hline
                                       & \textbf{BPIC 2012 Complete} & \textbf{BPIC 2012 A} & \textbf{BPIC 2012 O} & \textbf{BPIC 2012 W Complete} \\ \hline
\textit{Pasquadibisceglie et al.~\cite{pasquadibisceglie_using_2019}}      & 74.55                       & 71.47                & 77.51                & 66.14                         \\
\textit{Tax el al.~\cite{Tax2017}}                    & 79.39                       & 77.75                & 81.22                & 67.80                         \\
\textit{Camargo et al.~\cite{Camargo2019}}                & 79.22                       & 78.92                & 85.13                & 65.19                         \\
\textit{Hinkka el al.~\cite{hinkka_exploiting_2020}}                 & {79.76}                 & {79.27}          & {85.51}          & 67.24                         \\
\textit{Khan et al.~\cite{khan_memory-augmented_2018}}                   & 75.50                       & 75.62                & 84.48                & 75.91                         \\
\textit{Evermann et al.~\cite{Evermann2017}}               & 63.37                       & 74.44                & 79.20                & 65.38                         \\
\textit{Di Mauro et al.~\cite{di_mauro_activity_2019}}                  & 78.72                       & 78.09                & 81.52                & 65.01                         \\
\textit{Theis et al. (w/o attributes)~\cite{theis_decay_2019}} & 73.10                       & 66.23                & 81.52                & 76.97                         \\
\textit{Theis et al. (w/ attributes)~\cite{theis_decay_2019}}  & 65.21                       & 65.12                & 73.56                & 72.52                         \\
\textit{\textbf{Shared attention-based}}             & {\textbf{78.61}}            & \textbf{74.88}       & \textbf{82.05}       & {\textbf{84.39}}          \\
\textit{\textbf{Specialised attention-based}}              & \textbf{77.74}              & \textbf{74.86}       & \textbf{81.60}       & \textbf{83.75}                \\ \hline
\end{tabular}
}
\caption{Accuracy comparison in \textit{percentage} (\%)  w.r.t. benchmark results of Rama-Maneiro et al.~\cite{RamaManeiro2020}}
\label{table:benchmark}
\end{table}

\subsection{Analysis of Model Interpretations}
\label{subsec:interpretations-analysis}

\begin{figure}[htp!]
    \centering
    \includegraphics[width=0.8\textwidth]{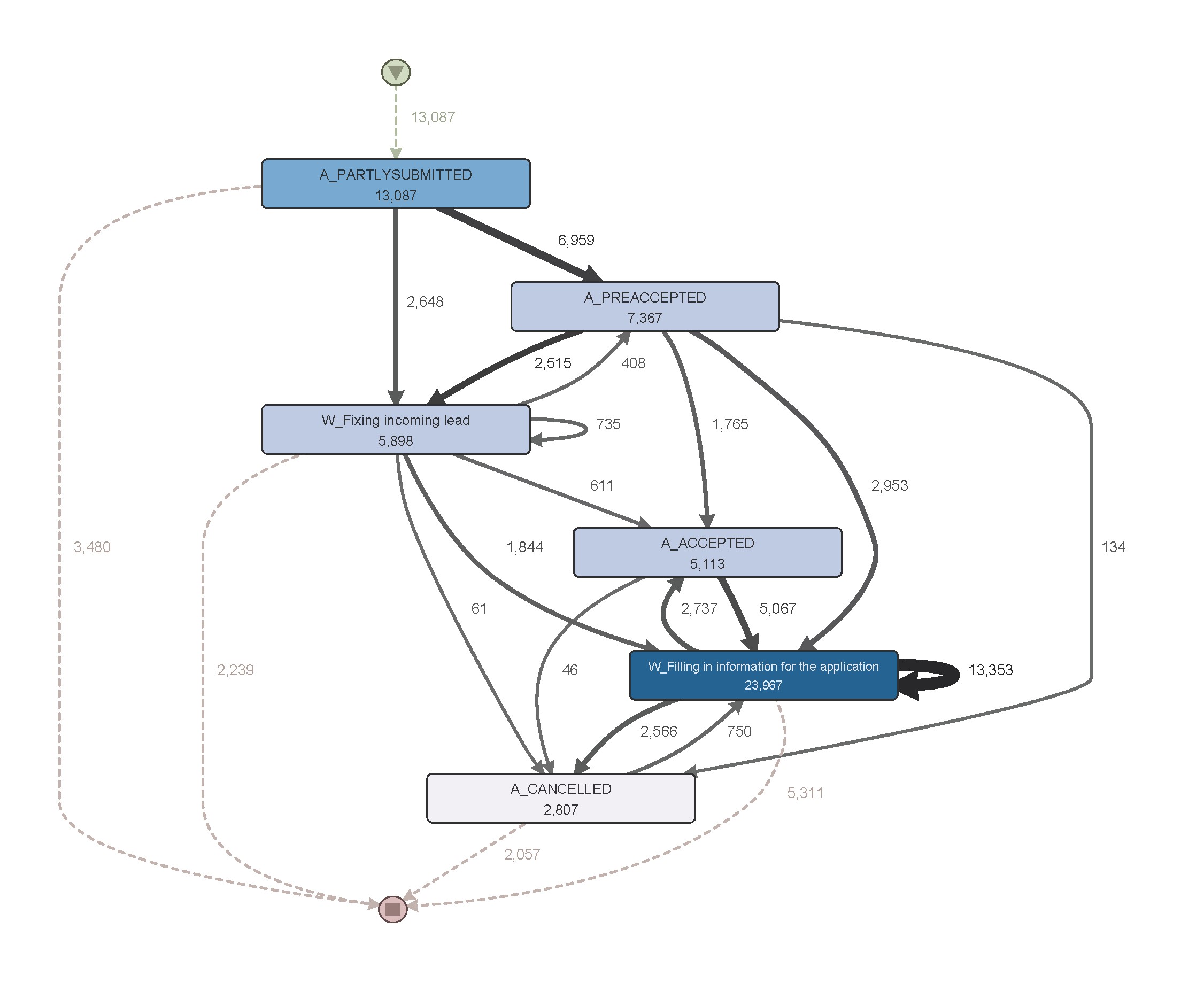}
    \caption{A partial process map for decision point~A\_PREACCEPTED.}
    \label{figure:decision_point_1}
\end{figure}

\begin{figure}[htp!]
    \centering
    \includegraphics[width=\textwidth]{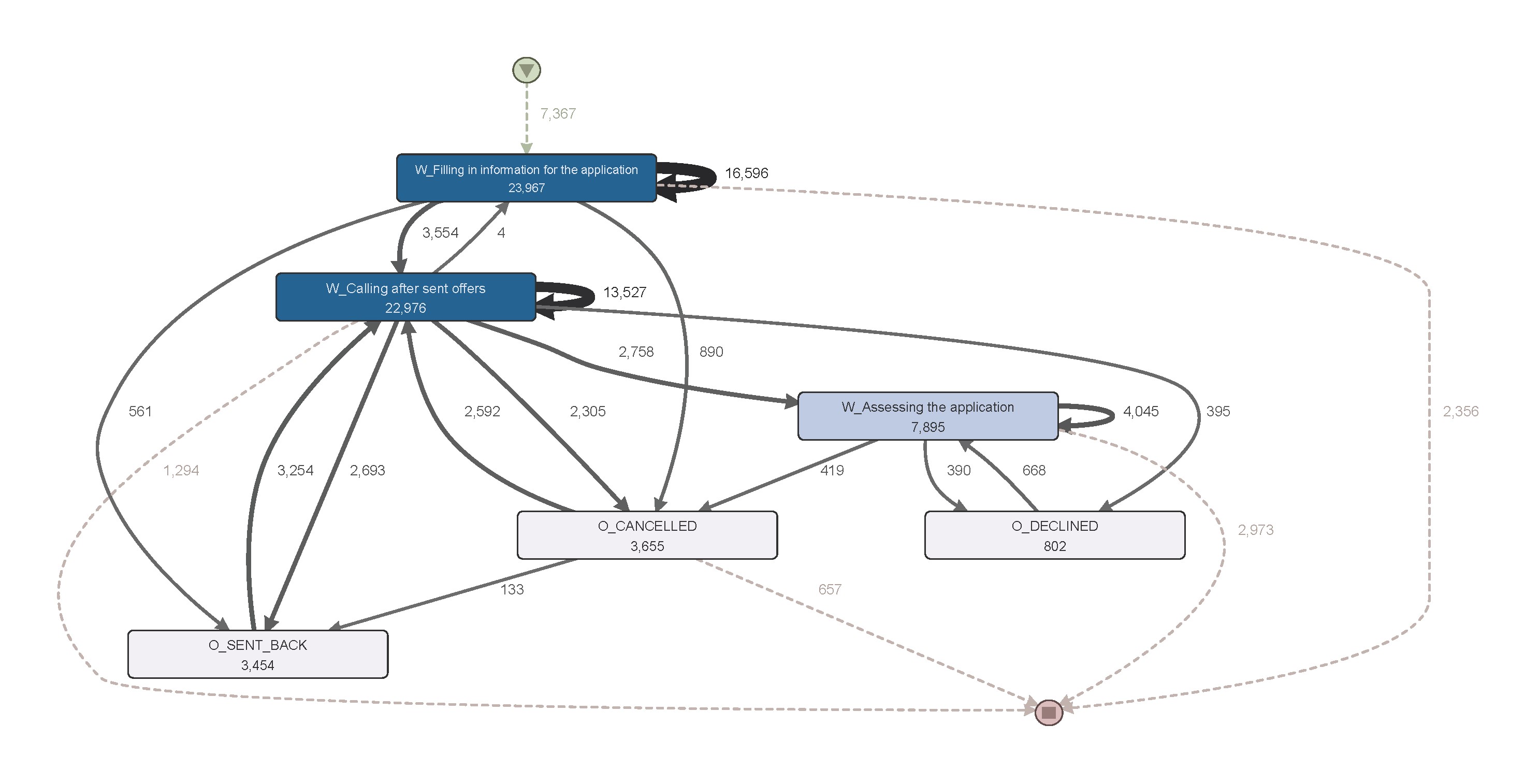}
    \caption{A partial process map for decision point~W\_Call for Sent Offers.}
    \label{figure:decision_point_2}
\end{figure}

We analyse the interpretations generated for the specialised attention-based model at two key decision points of the process, and we compare these against the shared attention-based model interpretations as well as the insights gathered from domain knowledge. 
We use process maps to visually present certain process knowledge relevant to the event log data (see Figure~\ref{figure:decision_point_1} and Figure~\ref{figure:decision_point_2}). The two decision points at which we extract interpretations are as follows. A\_PREACCEPTED, shown in Figure~\ref{figure:decision_point_1}, is the loan application pre-acceptance stage. After this decision point, the application can either get accepted (A\_ACCEPTED), cancelled or declined (A\_CANCELLED\_DECLINED), moved to fill out missing information of the application (W\_Filling in information for the application) or to fix issues of the incoming application (W\_Fixing incoming lead). The second decision point, shown in Figure~\ref{figure:decision_point_2}, is W\_Call for Sent Offers. This decision point occurs after the bank sends out an offer to the customer. 

\paragraph{\bf Interpretations for predictions at A\_PREACCEPTED}

At this decision point, both models predict that the next activity should either be W\_Filling in information for the application, or W\_Fixing incoming lead, possibly due to lack of sufficient data.

\begin{figure}[htp!]
    \includegraphics[width=\textwidth]{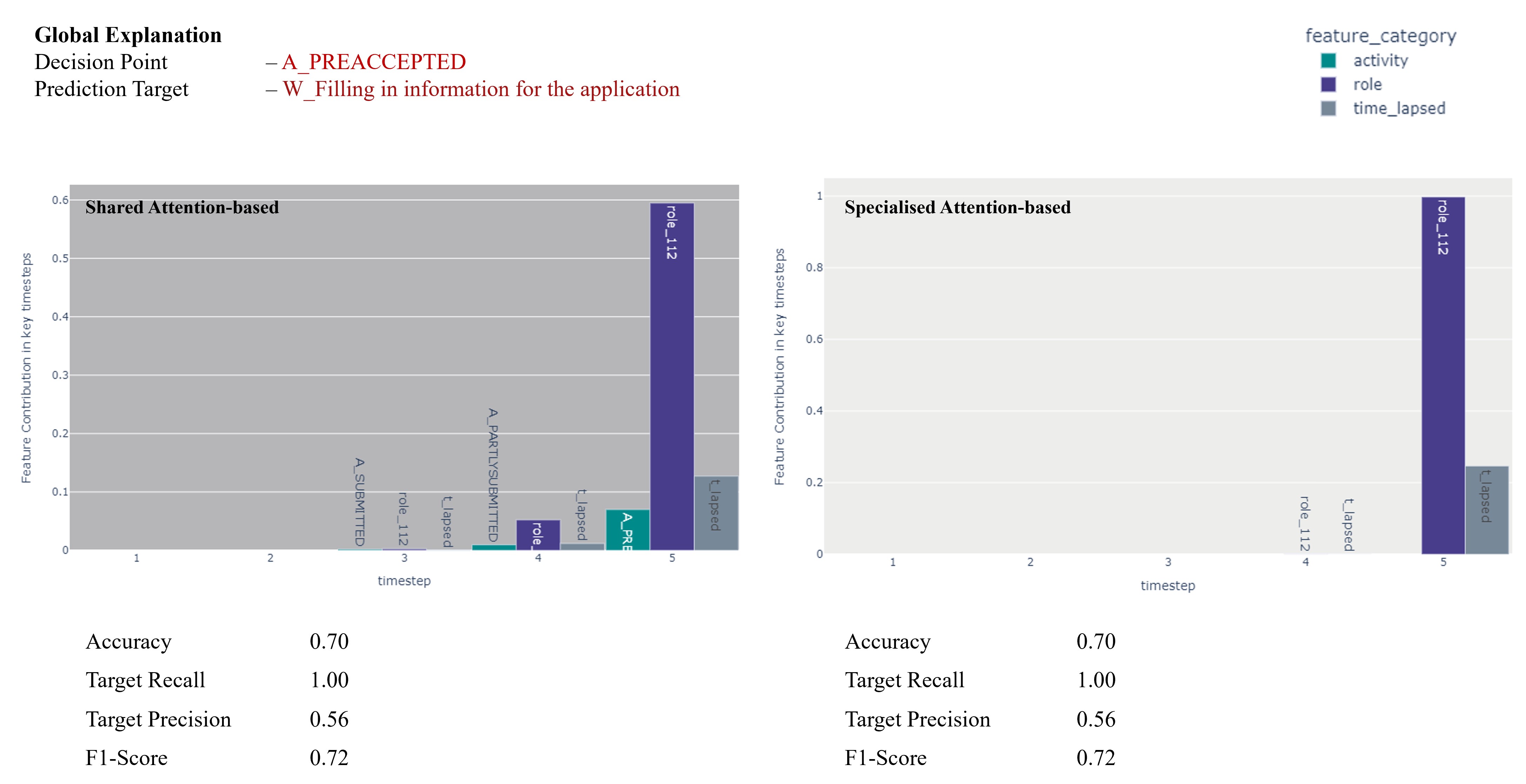}
    \caption{Global Explanation for `W\_Filling in information for the application' target at `A\_PREACCEPTED' decision point}
    \label{fig:Global_A_PRE_W_Filling}
\end{figure}

\begin{figure}[htp!]
    \includegraphics[width=\textwidth]{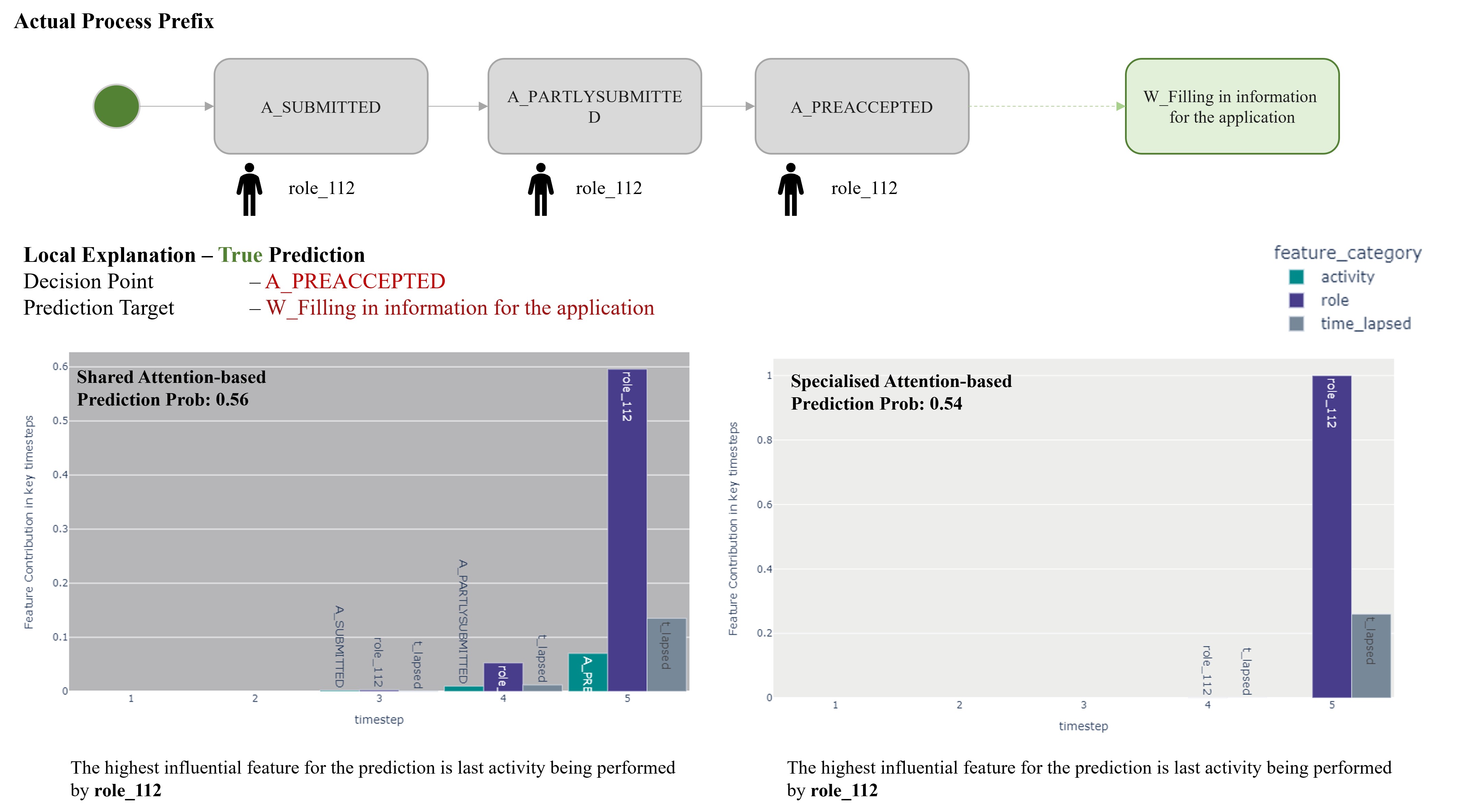}
    \caption{Local Explanation for a~\textbf{True} prediction for `W\_Filling in information for the application' target at `A\_PREACCEPTED' decision point}
    \label{fig:Local_A_PRE_W_Filling_True}
\end{figure}

\begin{figure}[htp!]
    \includegraphics[width=\textwidth]{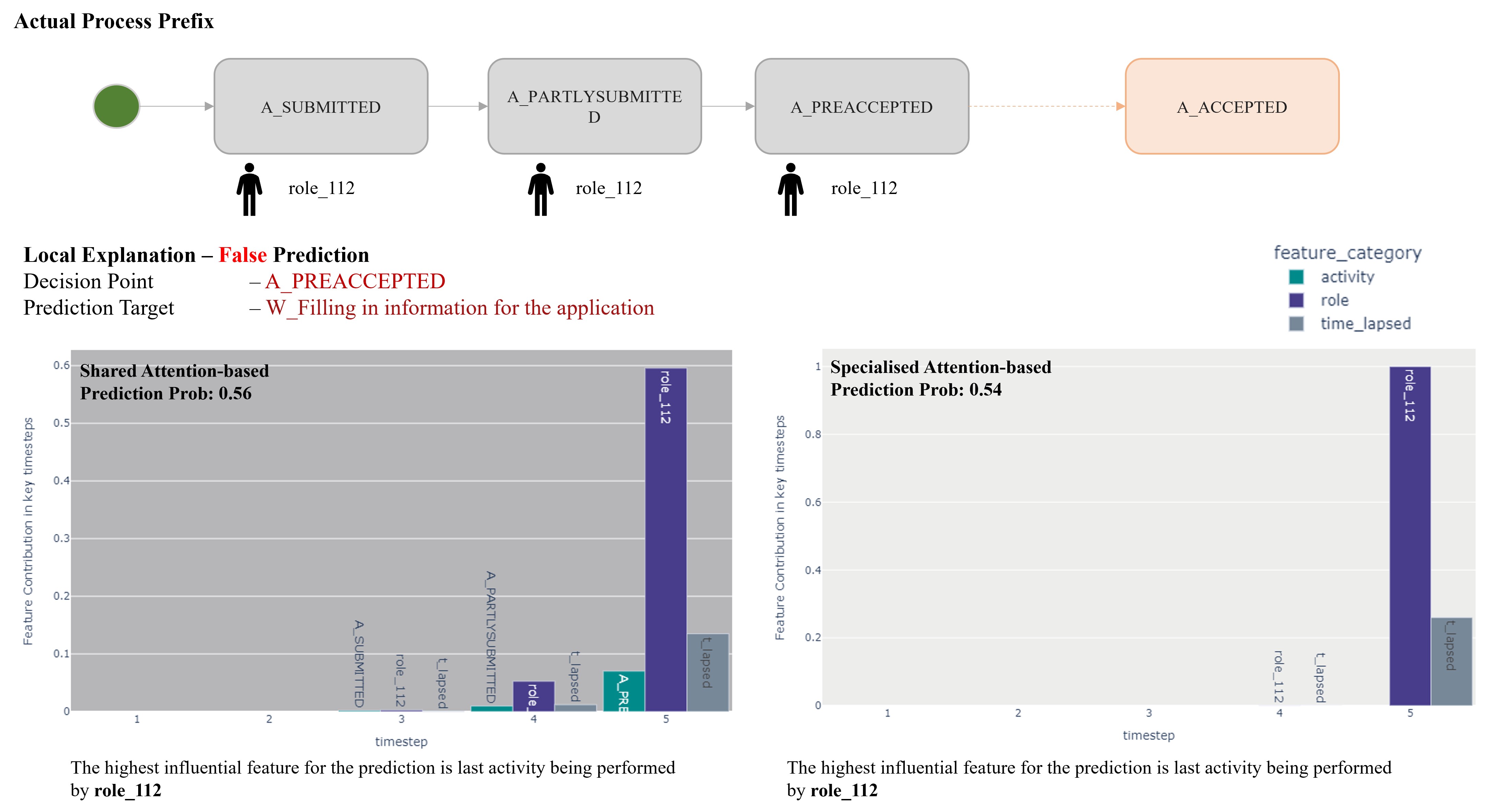}
    \caption{Local Explanation for a~\textbf{False} prediction for `W\_Filling in information for the application' target at `A\_PREACCEPTED' decision point}
    \label{fig:Local_A_PRE_W_Filling_False}
\end{figure}

Figure~\ref{fig:Global_A_PRE_W_Filling} shows the global interpretation for the prediction target (one of the possible next activities on the process trace that comes after the decision point denoted by the activity label`A\_PREACCEPTED') `W\_Filling in information for the application'. Both the model interpretations suggest that it pays a high attention to the feature value  of `role\_112' (which belongs to the feature category `role') having performed the last activity of the prefix. `role\_112' is one of the agents who perform one of the specific activities in the loan application process.The local interpretations generated for a True prediction (Figure~\ref{fig:Local_A_PRE_W_Filling_True}) and a False prediction (Figure~\ref{fig:Local_A_PRE_W_Filling_False}) also remain consistent to the relevant global interpretation. One thing to note is the prediction probability for this prediction is very weak for all the test cases, and does not exceed 0.55.

\begin{figure}[htp!]
    \includegraphics[width=\textwidth]{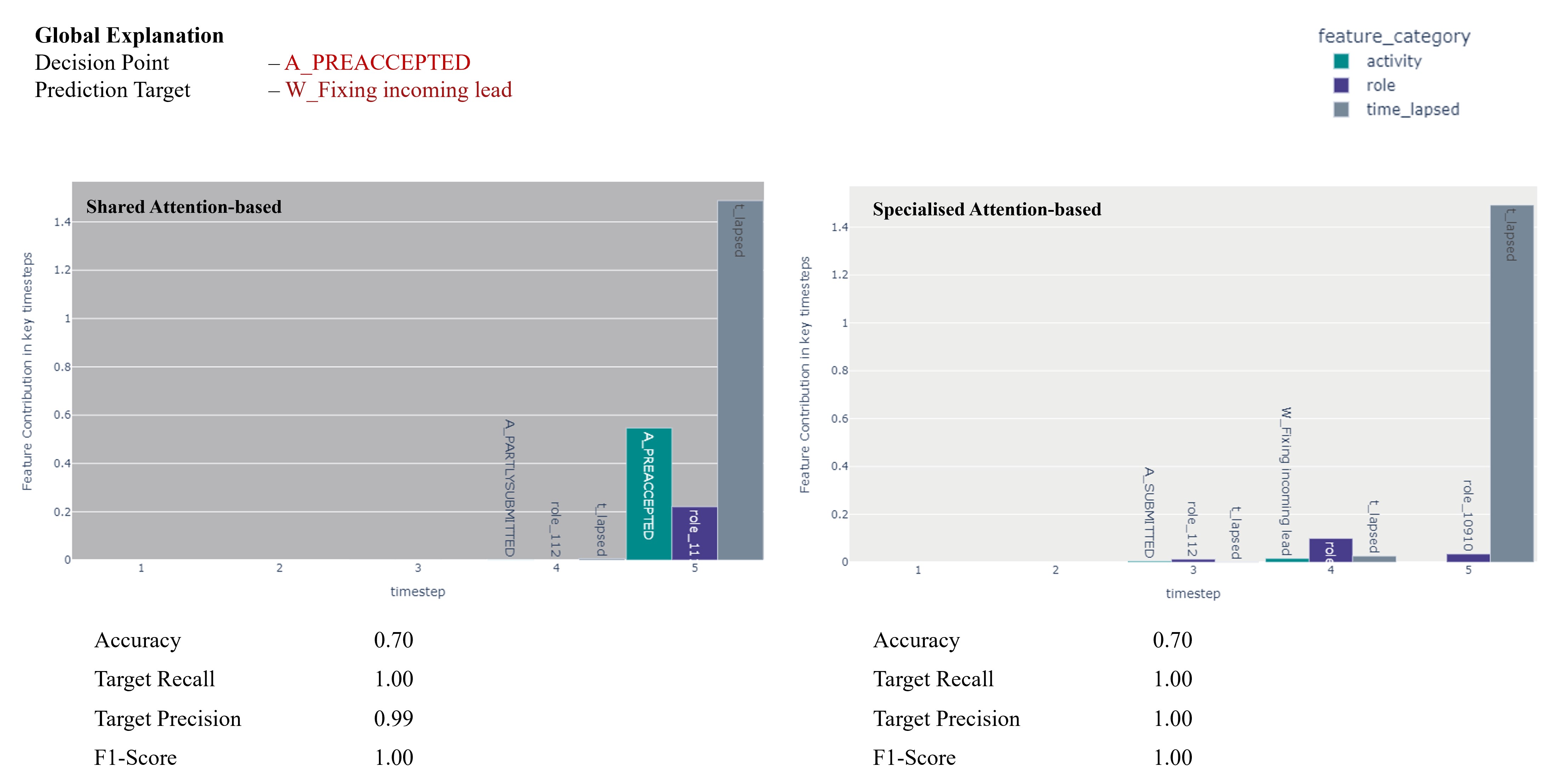}
    \caption{Global Explanation for `W\_Fixing incoming lead' target at `A\_PREACCEPTED' decision point}
    \label{fig:Global_A_PRE_W_Fixing}
\end{figure}

Figure~\ref{fig:Global_A_PRE_W_Fixing} shows the global interpretation for the prediction target `W\_Fixing incoming lead'. As per the interpretation from both models, when generating the decision, model pays a high attention to the feature `time lapsed'. This prediction target has a perfect precision and recall.

\textit{\textbf{Diagnosis}}: 
\highlight{Since the decision point A\_PREACCEPTED occurs much early in the process, the model does not have sufficient data to decide the next activity. The activity sequence up to that point is almost identical for all the traces. Hence it uses the minimum information it has to make the best guess. All the traces that does not continue to the next activity `W\_Fixing incoming lead' after the decision point A\_PREACCEPTED are performed by one resource, `role\_112' which is likely to be an automated agent. Hence, the model looks at this one feature value and makes a very weak guess that the next activity of those prefixes should be `W\_Filling in information for the application', which is the majority class in that group. However, the traces that go into the `W\_Fixing incoming lead' next activity are performed by a wide pool of resources. Hence, the model cannot pay attention to one particular resource when making the decision. Instead it pays attention to `time lapsed' feature to make a good prediction (Figure~\ref{fig:Global_A_PRE_W_Fixing}).}

\paragraph{\bf Interpretations for predictions at W\_Calling for sent offers}

This decision point occurs later in the loan application process, when the bank performs the initial screening of the accepted applications and sends an offer for eligible customers. Once the offer is sent (i.e., following the activity label O\_SENT in the process trace), bank will periodically followup on the offer until the customer acknowledges (Activity label O\_SENT\_BACK) the offer. The decision point denoted by the activity label `W\_Calling for sent offers' represents this followup phase. If the customer does not acknowledge the offer, the offer may get cancelled or declined (activity label `O\_CANCELLED\_DECLINED')~\cite{Bautista2013}.

\begin{figure}[htp!]
    \includegraphics[width=\textwidth]{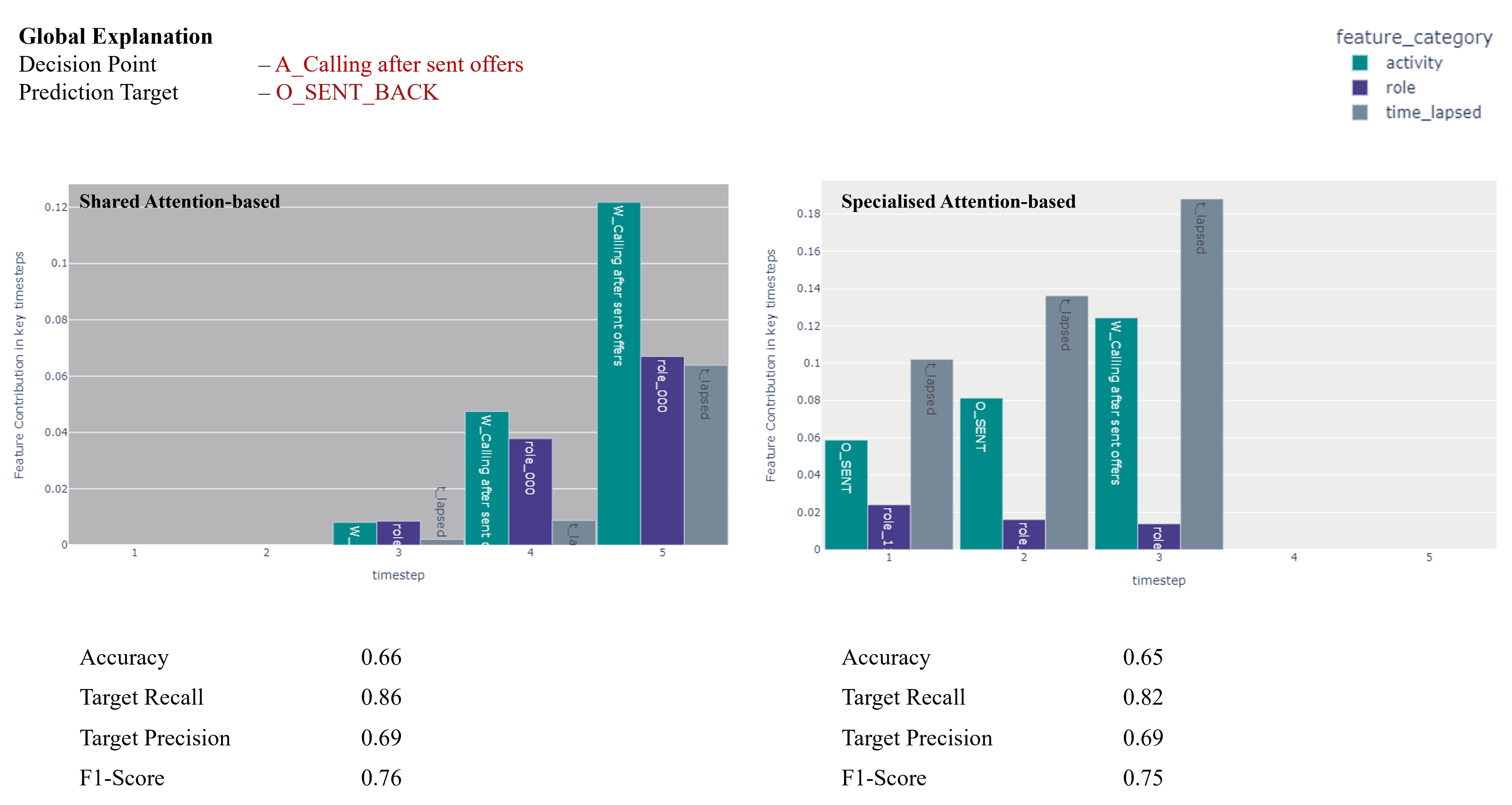}
    \caption{Global Explanation for `O\_SENT\_BACK' target at `W\_Calling for sent offers' decision point}
    \label{fig:Global_W_Call_O_SB}
\end{figure}

Figure~\ref{fig:Global_W_Call_O_SB} depicts the reasoning extracted from the shared and specialised attention-based mechanisms for a prediction of activity label `O\_SENT\_BACK'. Whilst the shared explanation indicates that the highest attention being placed on last 3 `W\_Calling for sent offers' activity labels occurred prior the decision point, specialised model explanation suggests that it pays the highest attention to the last occurred `O\_SENT' activity label, along with the activity label `W\_Calling for sent offers' and the time lapsed feature. 

\begin{figure}[htp!]
    \includegraphics[width=\textwidth]{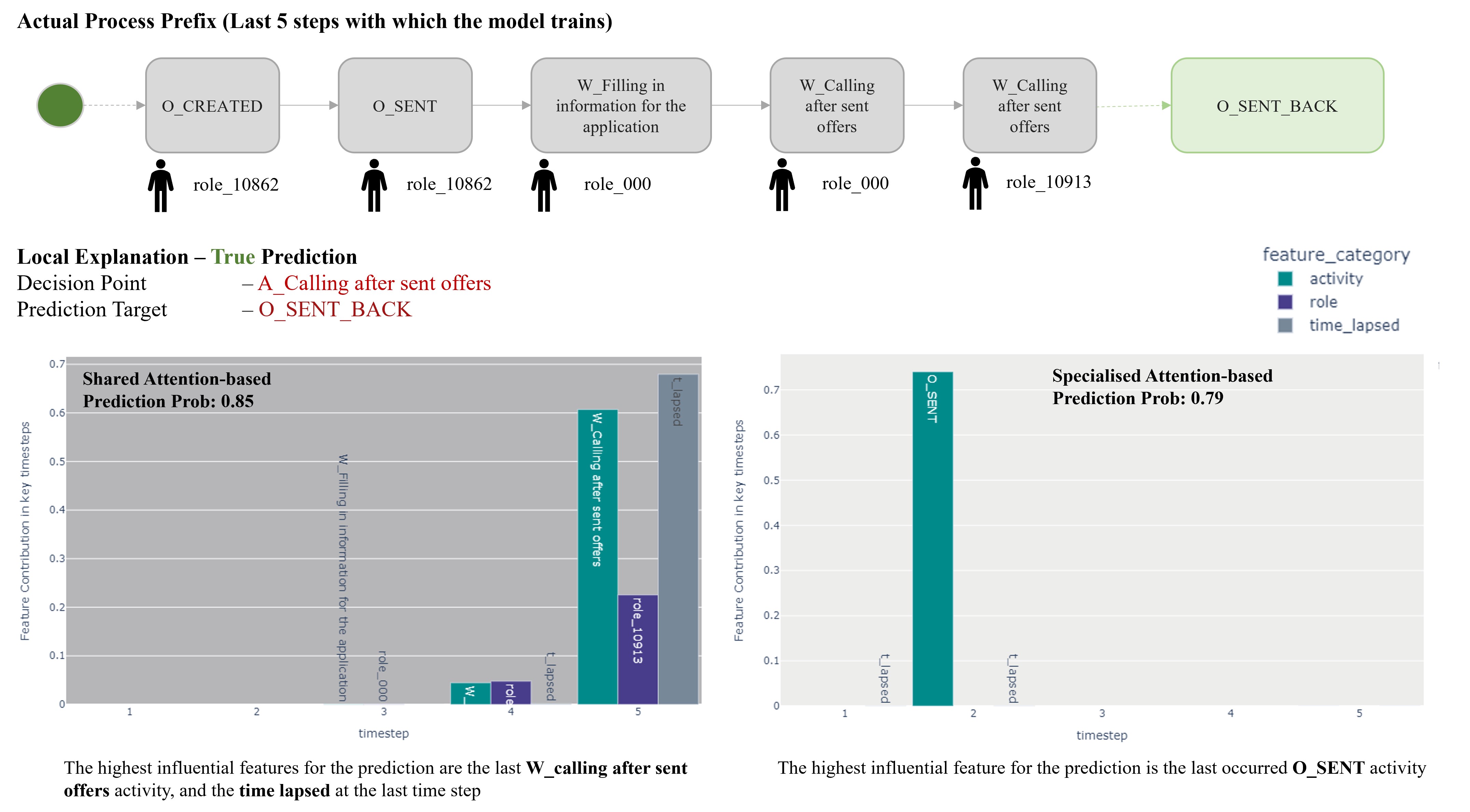}
    \caption{Local Explanation for a~\textbf{True} prediction for `O\_SENT\_BACK' target at `W\_Calling for sent offers' decision point}
    \label{fig:Local_W_Call_O_SB_True}
\end{figure}

\begin{figure}[htp!]
    \includegraphics[width=\textwidth]{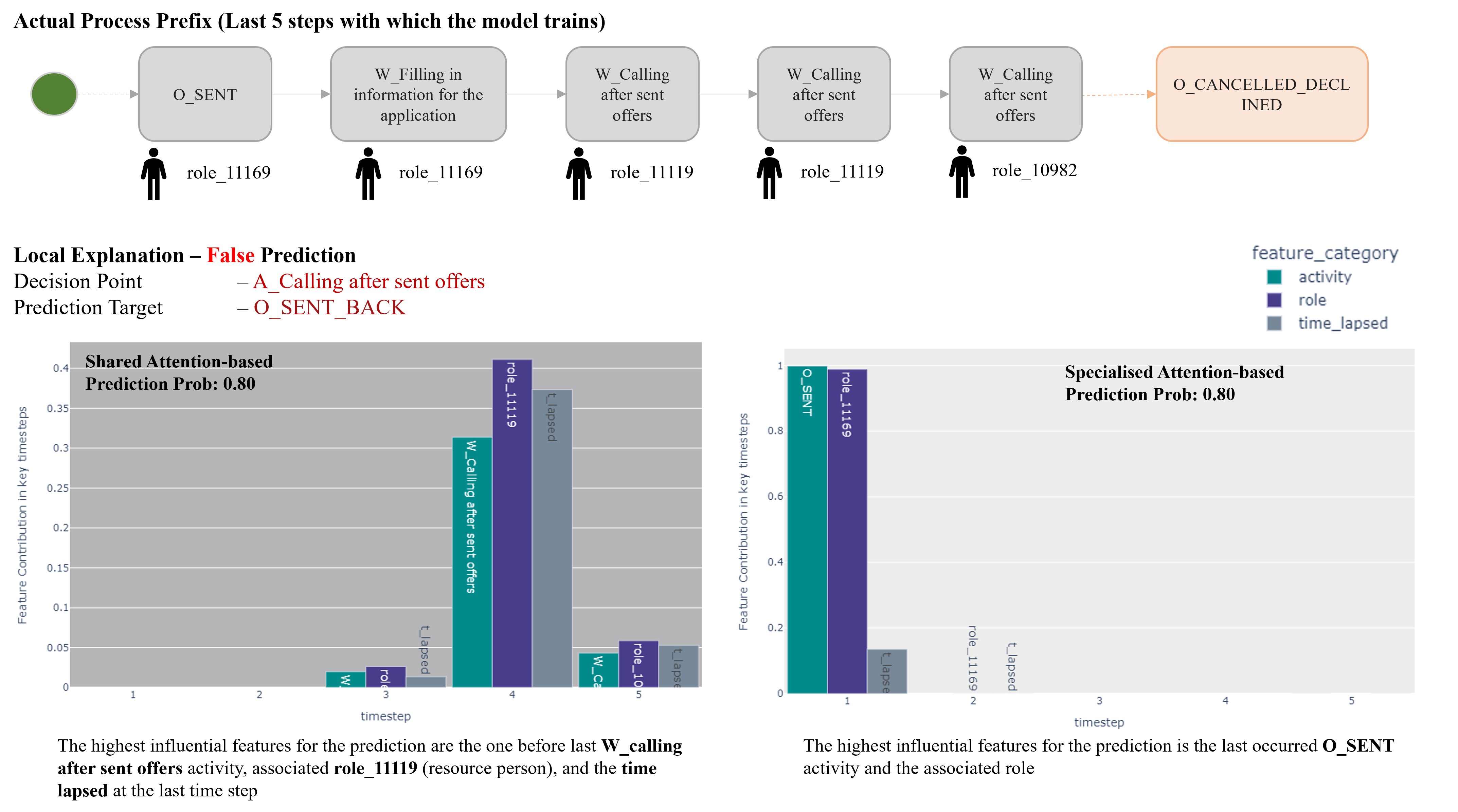}
    \caption{Local Explanation for a~\textbf{False} prediction for 'O\_SENT\_BACK' target at 'W\_Calling for sent offers' decision point}
    \label{fig:Local_W_Call_O_SB_False}
\end{figure}

At a local level, shown in Figure~\ref{fig:Local_W_Call_O_SB_True} and Figure~\ref{fig:Local_W_Call_O_SB_False}, it is further clear that the specialised attention-based model explanation is more focused on the last occurred `O\_SENT' activity, to predict an `O\_SENT\_BACK' next activity. This explanation makes more sense as per the actual process flow, compared to the explanations generated by the shared attention-based model. 

\begin{figure}[htp!]
    \includegraphics[width=\textwidth]{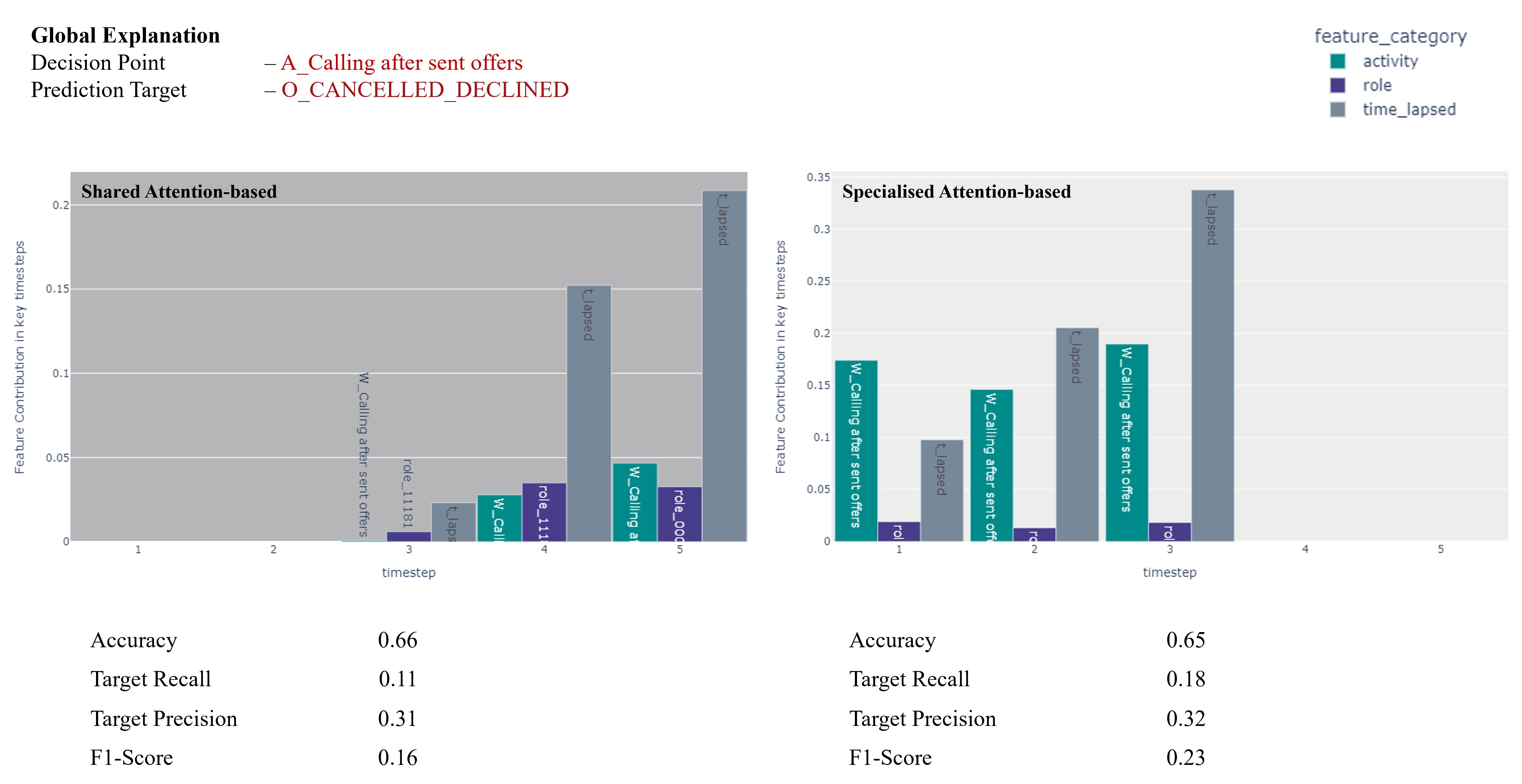}
    \caption{Global Explanation for 'O\_CANCELLED\_DECLINED' target at 'W\_Calling for sent offers' decision point}
    \label{fig:Global_W_Call_O_CD}
\end{figure}

Whereas, the global explanation behind the prediction of activity label 'O\_CANCELLED\_DECLINED' suggests that both the models pay a high attention to the `time lapsed' feature and multiple occurrences of the activity label 'W\_Calling for sent offers' closer to the decision point (Figure~\ref{fig:Global_W_Call_O_CD}). In the real process context, this indicates the cancellation of an application for which an acknowledgement not received from the customer despite multiple followups.

\subsection{Discussion}

In the comparison between  shared attention-based model and  specialised attention-based model, we can observe that the two models perform similar in terms of model performance. Shared attention-based model considers all three attributes of an event (activity, role and time) simultaneously when computing the attention. This mechanism is more aligned with the actual business process execution since these three variables are closely interlinked in defining a process event, hence may have a synergy effect on determining the next event. 

The specialised attention-based model is designed with a focus on interpretability, whilst ensuring that, in terms of model performance, it does not fall behind the shared attention-based model or the other comparable benchmarks. 
Our work adopts the following two key design decisions to facilitate better interpretability. 
First, one-hot encoding of categorical features is used to establish an orthogonal relationship between the feature values that belong to the same feature group. 
Second, independent attribute attention mechanism is designed for the model to pick the most important attribute of a given event, instead of classifying all three attributes of a given high important event being important. 
We can observe the strengths of these `interpretability enhancement' design in certain interpretations, against the shared attention-based mechanism in Section~\ref{subsec:interpretations-analysis}.
\section{Conclusion and Future Work}
\label{sec:concl}

We have presented two different attention-based LSTM architectures to predict the next activity along a process execution, considering the event attributes of activity, resource and timestamp. The two attention mechanisms have been designed with the primary objective of extracting the interpretation of the decision making logic of each model. The shared attention-based model considers all three event attributes together to make the decision, whereas the specialised attention-based model focuses on extracting independent contributions of the input event attributes. The two approaches agree as well as disagree with each other on model interpretations for different cases. By analysing interpretations of predictions made at process decision points, we can observe that specialised attention-based approach often gives better interpretations --- they make more sense as validated against the relevant domain knowledge.

In this work, we have limited our prediction target on the next activity label, and considered only three event(-related) attributes as model features. 
In future work, we expect to expand this model to predict not only the next activity label but the resource and timestamp of the next activity, to extend the current approach for process outcome prediction and remaining time prediction, and to extract the model interpretations for each of the prediction targets. 
We are also interested in using the generated explanations to diagnose and improve the existing model, e.g., to inspect the prediction instances of the model in which it under-performs. To this end, we have started to build a framework for generating and presenting purpose-driven explanations~\cite{CAiSE-forum-2022}. 
Last but not least, in regard to the underlying models we plan to improve the predictive performance of the proposed architecture by optimising the hyperparmeters~\cite{Kaselimi2019} and explore any possible relationship between the model hyperparameters and model interpretations. 



\paragraph{\bf Acknowledgements} 
The reported research is part of a PhD project supported by a Science and Engineering Faculty scholarship and a Centre for Data Science top up scholarship at Queensland University of Technology (QUT). It also received funding support from Centre for Data Science First Byte Funding Program at QUT as well as QUT's Women in Research Grant Scheme.


\paragraph{\bf Reproducibility} 
The source code for experiment can be downloaded from \href{https://git.io/JE9fT}{\textbf{\textit{https://git.io/JE9fT}}}.

\bibliography{Ref}

\begin{thebibliography}{10}
\expandafter\ifx\csname url\endcsname\relax
  \def\url#1{\texttt{#1}}\fi
\expandafter\ifx\csname urlprefix\endcsname\relax\def\urlprefix{URL }\fi
\expandafter\ifx\csname href\endcsname\relax
  \def\href#1#2{#2} \def\path#1{#1}\fi

\bibitem{guidotti2018}
R.~Guidotti, A.~Monreale, S.~Ruggieri, F.~Turini, F.~Giannotti, D.~Pedreschi, A
  survey of methods for explaining black box models, ACM Comput. Surv. 51~(5)
  (2018) 93:1--93:42.

\bibitem{Evermann2017}
J.~Evermann, J.~Rehse, P.~Fettke, Predicting process behaviour using deep
  learning, Decis. Support Syst. 100 (2017) 129--140.

\bibitem{Tax2017}
N.~Tax, I.~Verenich, M.~L. Rosa, M.~Dumas, Predictive business process
  monitoring with {LSTM} neural networks, in: Advanced Information Systems
  Engineering, Vol. 10253 of Lecture Notes in Computer Science, Springer, 2017,
  pp. 477--492.

\bibitem{Camargo2019}
M.~Camargo, M.~Dumas, O.~G. Rojas, Learning accurate {LSTM} models of business
  processes, in: International Conference on Business Process Management, Vol.
  11675 of Lecture Notes in Computer Science, Springer, 2019, pp. 286--302.

\bibitem{verenich2019survey}
I.~Verenich, M.~Dumas, M.~L. Rosa, F.~M. Maggi, I.~Teinemaa, Survey and
  cross-benchmark comparison of remaining time prediction methods in business
  process monitoring, {ACM} Trans. Intell. Syst. Technol. 10~(4) (2019)
  34:1--34:34.

\bibitem{Galanti2020}
R.~Galanti, B.~Coma{-}Puig, M.~de~Leoni, J.~Carmona, N.~Navarin, Explainable
  predictive process monitoring, in: 2020 2nd International Conference on
  Process Mining (ICPM), {IEEE}, 2020, pp. 1--8.

\bibitem{Sindhgatta2020b}
R.~Sindhgatta, C.~Ouyang, C.~Moreira, Exploring interpretability for predictive
  process analytics, in: International Conference on Service-Oriented
  Computing, Vol. 12571 of Lecture Notes in Computer Science, Springer, 2020,
  pp. 439--447.

\bibitem{serrano2019attention}
S.~Serrano, N.~A. Smith, Is attention interpretable?, in: Proc. of the 57th
  Conference of the Association for Computational Linguistics, {ACL},
  Association for Computational Linguistics, 2019, pp. 2931--2951.

\bibitem{Retain_Choi2016}
E.~Choi, M.~T. Bahadori, J.~Sun, J.~Kulas, A.~Schuetz, W.~F. Stewart, {RETAIN:
  A}n interpretable predictive model for healthcare using reverse time
  attention mechanism, in: Proceedings of the 30th International Conference on
  Neural Information Processing Systems, 2016, pp. 3512--–3520.

\bibitem{Sindhgatta2020a}
R.~Sindhgatta, C.~Moreira, C.~Ouyang, A.~Barros, Exploring interpretable
  predictive models for business processes, in: International Conference on
  Business Process Management, Vol. 12168 of Lecture Notes in Computer Science,
  Springer, 2020, pp. 257--272.

\bibitem{Dumas_La_2013}
M.~Dumas, M.~L. Rosa, J.~Mendling, H.~A. Reijers, Fundamentals of Business
  Process Management, Second Edition, Springer, 2018.

\bibitem{van_der_aalst_process_2016}
W.~M.~P. {van der Aalst}, Process Mining: Data Science in Action, Second
  Edition, Springer, 2016.

\bibitem{BPIC2012}
B.~van Dongen, {BPI Challenge 2012. 4TU.ResearchData. Dataset} (2012).
\newblock \href
  {http://dx.doi.org/10.4121/uuid:3926db30-f712-4394-aebc-75976070e91f}
  {\path{doi:10.4121/uuid:3926db30-f712-4394-aebc-75976070e91f}}.

\bibitem{xes_ieee_2016}
{IEEE} standard for {eXtensible} event stream ({XES}) for achieving
  interoperability in event logs and event streams, {IEEE} Std 1849-2016.

\bibitem{teinemaa2019outcome}
I.~Teinemaa, M.~Dumas, M.~L. Rosa, F.~M. Maggi, Outcome-oriented predictive
  process monitoring: Review and benchmark, {ACM} Trans. Knowl. Discov. Data
  13~(2) (2019) 17:1--17:57.

\bibitem{Samek2017}
W.~Samek, T.~Wiegand, K.-R. Müller, Explainable artificial intelligence:
  Understanding, visualizing and interpreting deep learning models (2017).
\newblock \href {http://arxiv.org/abs/1708.08296} {\path{arXiv:1708.08296}}.

\bibitem{Ribeiro2016}
M.~T. Ribeiro, S.~Singh, C.~Guestrin, {``Why Should I Trust You?'': Explaining
  the Predictions of Any Classifier}, in: Proceedings of the 22nd {ACM}
  {SIGKDD} International Conference on Knowledge Discovery and Data Mining, San
  Franciso, California, 2016.

\bibitem{Lundberg2017}
S.~M. Lundberg, S.-I. Lee, A unified approach to interpreting model
  predictions, in: Proceedings of the 2017 Neural Jnformation Processing
  Systems Conference, Long Beach, USA, 2017.

\bibitem{Giri2020}
T.~De, P.~Giri, A.~Mevawala, R.~Nemani, A.~Deo, Explainable {AI}: A hybrid
  approach to generate human-interpretable explanation for deep learning
  prediction, Procedia Comput. Sci. 168 (2020) 40--48.

\bibitem{Boz2002}
O.~Boz, Extracting decision trees from trained neural networks, in: Proceedings
  of the Eighth {ACM} {SIGKDD} International Conference on Knowledge Discovery
  and Data Mining, {ACM}, 2002, pp. 456--461.

\bibitem{Wol2020}
A.~Wolanin, G.~Mateo-Garc{\'{\i}}a, G.~Camps-Valls, L.~G{\'{o}}mez-Chova,
  M.~Meroni, G.~Duveiller, Y.~Liangzhi, L.~Guanter, Estimating and
  understanding crop yields with explainable deep learning in the indian wheat
  belt, Environ. Res. Lett. 15~(2) (2020) 024019.

\bibitem{Iad2021}
G.~Iadarola, F.~Martinelli, F.~Mercaldo, A.~Santone, Towards an interpretable
  deep learning model for mobile malware detection and family identification,
  Comput. Secur. 105 (2021) 102198.

\bibitem{Mehdiyev2020}
N.~Mehdiyev, P.~Fettke, Prescriptive process analytics with deep learning and
  explainable artificial intelligence, in: Proceedings of the 28th European
  Conference on Information Systems (ECIS), 2020, pp. 1--17.

\bibitem{Wein2020}
S.~Weinzierl, S.~Zilker, J.~Brunk, K.~Revoredo, M.~Matzner, J.~Becker, {XNAP:
  Making LSTM}-based next activity predictions explainable by using {LRP}, in:
  Business Process Management Workshops, Vol. 397 of Lecture Notes in Business
  Information Processing, Springer, 2020, pp. 129--141.

\bibitem{Xue2019}
Q.~Xue, M.~C. Chuah, Explainable deep learning based medical diagnostic system,
  Smart Health 13 (2019) 100068.

\bibitem{Retainvis_Kwon2019}
B.~C. Kwon, M.~Choi, J.~T. Kim, E.~Choi, Y.~B. Kim, S.~Kwon, J.~Sun, J.~Choo,
  Retainvis: Visual analytics with interpretable and interactive recurrent
  neural networks on electronic medical records, {IEEE} Trans. Vis. Comput.
  Graph. 25~(1) (2019) 299--309.

\bibitem{Velmurugan2021}
M.~Velmurugan, C.~Ouyang, C.~Moreira, R.~Sindhgatta, Evaluating fidelity of
  explainable methods for predictive process analytics, in: S.~Nurcan,
  A.~Korthaus (Eds.), Intelligent Information Systems - CAiSE Forum 2021,
  Melbourne, VIC, Australia, June 28 - July 2, 2021, Proceedings, Vol. 424 of
  Lecture Notes in Business Information Processing, Springer, 2021, pp. 64--72.

\bibitem{XAIQuality_Zhou2021}
J.~Zhou, A.~H. Gandomi, F.~Chen, A.~Holzinger, Evaluating the quality of
  machine learning explanations: A survey on methods and metrics, Electronics
  10~(5) (2021) 593.

\bibitem{Evermann2017_2}
J.~Evermann, J.~Rehse, P.~Fettke, A deep learning approach for predicting
  process behaviour at runtime, in: International Conference on Business
  Process Management, Vol. 281 of Lecture Notes in Business Information
  Processing, 2016, pp. 327--338.

\bibitem{Park2020}
G.~Park, M.~Song, Predicting performances in business processes using deep
  neural networks, Decis. Support Syst. 129.

\bibitem{di_mauro_activity_2019}
N.~{Di Mauro}, A.~Appice, T.~M.~A. Basile, Activity prediction of business
  process instances with inception {CNN} models, in: AI*IA 2019 -- Advances in
  Artificial Intelligence, Vol. 11946 of Lecture Notes in Computer Science,
  Springer, 2019, pp. 348--361.

\bibitem{pasquadibisceglie_using_2019}
V.~Pasquadibisceglie, A.~Appice, G.~Castellano, D.~Malerba, Using convolutional
  neural networks for predictive process analytics, in: 2019 International
  Conference on Process Mining (ICPM), {IEEE}, 2019, pp. 129--136.

\bibitem{bukhsh2021processtransformer}
Z.~A. Bukhsh, A.~Saeed, R.~M. Dijkman, Process{T}ransformer: Predictive
  business process monitoring with transformer network (2021).
\newblock \href {http://arxiv.org/abs/2104.00721} {\path{arXiv:2104.00721}}.

\bibitem{Harl2020}
M.~Harl, S.~Weinzierl, M.~Stierle, M.~Matzner, Explainable predictive business
  process monitoring using gated graph neural networks, J. Decis. Syst. (2020)
  1--16.

\bibitem{Mehdiyev2021}
N.~Mehdiyev, P.~Fettke, Explainable artificial intelligence for process mining:
  A general overview and application of a novel local explanation approach for
  predictive process monitoring, in: Interpretable Artificial Intelligence: A
  Perspective of Granular Computing, Vol. 937, Springer Nature, 2021, pp.
  1--28.

\bibitem{Suriadi2015}
S.~Suriadi, C.~Ouyang, W.~M.~P. van~der Aalst, A.~H.~M. ter Hofstede, Event
  interval analysis: Why do processes take time?, Decis. Support Syst. 79
  (2015) 77--98.

\bibitem{Dongen2017}
B.~van Dongen, {BPI Challenge 2017. 4TU.ResearchData. Dataset} (2017).
\newblock \href
  {http://dx.doi.org/10.4121/uuid:5f3067df-f10b-45da-b98b-86ae4c7a310b}
  {\path{doi:10.4121/uuid:5f3067df-f10b-45da-b98b-86ae4c7a310b}}.

\bibitem{Kaselimi2019}
M.~Kaselimi, N.~Doulamis, A.~D. Doulamis, A.~Voulodimos, E.~Protopapadakis,
  Bayesian-optimized bidirectional {LSTM} regression model for non-intrusive
  load monitoring, in: {IEEE} International Conference on Acoustics, Speech and
  Signal Processing, {ICASSP} 2019, Brighton, United Kingdom, May 12-17, 2019,
  {IEEE}, 2019, pp. 2747--2751.

\bibitem{RamaManeiro2020}
E.~Rama-Maneiro, J.~C. Vidal, M.~Lama, Deep learning for predictive business
  process monitoring: Review and benchmark (2021).
\newblock \href {http://arxiv.org/abs/2009.13251} {\path{arXiv:2009.13251}}.

\bibitem{SOKOLOVA_2009}
M.~Sokolova, G.~Lapalme, A systematic analysis of performance measures for
  classification tasks, Information Processing and Management 45~(4) (2009)
  427--437.

\bibitem{Bautista2013}
A.~D. Bautista, L.~Wangikar, S.~M.~K. Akbar, Process mining-driven optimization
  of a consumer loan approvals process -- the {BPIC} 2012 challenge case study,
  in: International Conference on Business Process Management, Vol. 132 of
  Lecture Notes in Business Information Processing, Springer, 2012, pp.
  219--220.

\bibitem{hinkka_exploiting_2020}
M.~Hinkka, T.~Lehto, K.~Heljanko, Exploiting event log event attributes in
  {RNN} based prediction, in: Data-Driven Process Discovery and Analysis, Vol.
  379 of Lecture Notes in Business Information Processing, Springer, 2019, pp.
  67--85.

\bibitem{khan_memory-augmented_2018}
A.~Khan, H.~Le, K.~Do, T.~Tran, A.~Ghose, H.~Dam, R.~Sindhgatta,
  Memory-augmented neural networks for predictive process analytics (2018).
\newblock \href {http://arxiv.org/abs/1802.00938} {\path{arXiv:1802.00938}}.

\bibitem{theis_decay_2019}
J.~Theis, H.~Darabi, Decay replay mining to predict next process events, {IEEE}
  Access 7 (2019) 119787--119803.

\bibitem{CAiSE-forum-2022}
B.~Wickramanayake, C.~Ouyang, C.~Moreira, Y.~Xu, Generating purpose-driven
  explanations: The case of process predictive model inspection, in:
  International Conference on Advanced Information Systems Engineering Forum
  (In Press), Springer, 2022.

\end{thebibliography}

\newpage

\section*{Appendix}

\subsection*{A.1. Significance test results for the performance analysis}

The ANOVA tests are run on the test sets of all four datasets to compare the differences between the shared and specialised models. According to Table~\ref{tab:st1}, Table~\ref{tab:st2}, Table~\ref{tab:st3} and Table~\ref{tab:st4}, all four tests report $p>.05$ and they are not statistically significant. We can suppose that shared and specialised attention models do not differ much.

\begin{table}[h]
\begin{tabular}{lllllll}
SUMMARY &  &  &  &  &  &  \\ \cline{1-5}
\multicolumn{1}{c}{\textit{Group}} & \multicolumn{1}{c}{\textit{Count}} & \multicolumn{1}{c}{\textit{Sum}} & \multicolumn{1}{c}{\textit{Average}} & \multicolumn{1}{c}{\textit{Variance}} &  &  \\ \cline{1-5}
Shared & \multicolumn{1}{r}{17277} & \multicolumn{1}{r}{257684} & \multicolumn{1}{r}{14.91486} & \multicolumn{1}{r}{31.97845} &  &  \\
Specialised & \multicolumn{1}{r}{17277} & \multicolumn{1}{r}{256137} & \multicolumn{1}{r}{14.82532} & \multicolumn{1}{r}{33.50063} &  &  \\ \cline{1-5}
 &  &  &  &  &  &  \\
ANOVA &  &  &  &  &  &  \\ \hline
\multicolumn{1}{c}{\textit{Source of Variation}} & \multicolumn{1}{c}{\textit{SS}} & \multicolumn{1}{c}{\textit{df}} & \multicolumn{1}{c}{\textit{MS}} & \multicolumn{1}{c}{\textit{F}} & \multicolumn{1}{c}{\textit{P-value}} & \multicolumn{1}{c}{\textit{F crit}} \\ \hline
Between Groups & \multicolumn{1}{r}{69.25997} & \multicolumn{1}{r}{1} & \multicolumn{1}{r}{69.25997} & \multicolumn{1}{r}{2.11548} & \multicolumn{1}{r}{ \textit{\textbf{0.14582}}} & \multicolumn{1}{r}{3.84173} \\
Within Groups & \multicolumn{1}{r}{1131217} & \multicolumn{1}{r}{34552} & \multicolumn{1}{r}{32.73954} & \multicolumn{1}{r}{} & \multicolumn{1}{r}{} & \multicolumn{1}{r}{} \\ \hline
 &  &  &  &  &  & 
\end{tabular}
\caption{Significant test results on the test split of BPIC 2012 Complete.}
\label{tab:st1}
\end{table}

\begin{table}[h]
\begin{tabular}{lllllll}
SUMMARY &  &  &  &  &  &  \\ \cline{1-5}
\multicolumn{1}{c}{\textit{Groups}} & \multicolumn{1}{c}{\textit{Count}} & \multicolumn{1}{c}{\textit{Sum}} & \multicolumn{1}{c}{\textit{Average}} & \multicolumn{1}{c}{\textit{Variance}} &  &  \\ \cline{1-5}
Shared & \multicolumn{1}{r}{5430} & \multicolumn{1}{r}{25773} & \multicolumn{1}{r}{4.74641} & \multicolumn{1}{r}{6.58092} &  &  \\
Specialised & \multicolumn{1}{r}{5430} & \multicolumn{1}{r}{25735} & \multicolumn{1}{r}{4.73941} & \multicolumn{1}{r}{6.34855} &  &  \\ \cline{1-5}
 &  &  &  &  &  &  \\
ANOVA &  &  &  &  &  &  \\ \hline
\multicolumn{1}{c}{\textit{Source of Variation}} & \multicolumn{1}{c}{\textit{SS}} & \multicolumn{1}{c}{\textit{df}} & \multicolumn{1}{c}{\textit{MS}} & \multicolumn{1}{c}{\textit{F}} & \multicolumn{1}{c}{\textit{P-value}} & \multicolumn{1}{c}{\textit{F crit}} \\ \hline
Between Groups & \multicolumn{1}{r}{0.13297} & \multicolumn{1}{r}{1} & \multicolumn{1}{r}{0.13297} & \multicolumn{1}{r}{0.02057} & \multicolumn{1}{r}{ \textit{\textbf{0.88597}}} & \multicolumn{1}{r}{3.84232} \\
Within Groups & \multicolumn{1}{r}{70194.07} & \multicolumn{1}{r}{10858} & \multicolumn{1}{r}{6.46473} & \multicolumn{1}{r}{} & \multicolumn{1}{r}{} & \multicolumn{1}{r}{} \\ \hline
 &  &  &  &  &  & 
\end{tabular}
\caption{Significant test results on the test split of BPIC 2012 A.}
\label{tab:st2}
\end{table}

\begin{table}[h]
\begin{tabular}{lllllll}
SUMMARY &  &  &  &  &  &  \\ \cline{1-5}
\multicolumn{1}{c}{\textit{Groups}} & \multicolumn{1}{c}{\textit{Count}} & \multicolumn{1}{c}{\textit{Sum}} & \multicolumn{1}{c}{\textit{Average}} & \multicolumn{1}{c}{\textit{Variance}} &  &  \\ \cline{1-5}
Shared & 7842 & 24220 & 3.08850 & 3.05045 &  &  \\
Specialised & 7842 & 24579 & 3.13428 & 3.03859 &  &  \\ \cline{1-5}
 &  &  &  &  &  &  \\
ANOVA &  &  &  &  &  &  \\ \hline
\multicolumn{1}{c}{\textit{Source of Variation}} & \multicolumn{1}{c}{\textit{SS}} & \multicolumn{1}{c}{\textit{df}} & \multicolumn{1}{c}{\textit{MS}} & \multicolumn{1}{c}{\textit{F}} & \multicolumn{1}{c}{\textit{P-value}} & \multicolumn{1}{c}{\textit{F crit}} \\ \hline
Between Groups & 8.21736 & 1 & 8.21736 & 2.69906 & \textit{\textbf{0.10043}} & 3.84205 \\
Within Groups & 47744.19 & 15682 & 3.04452 &  &  &  \\ \hline
 &  &  &  &  &  & 
\end{tabular}
\caption{Significant test results on the test split of BPIC 2012 O.}
\label{tab:st3}
\end{table}

\makeatletter
\setlength\@fptop{0pt} 
\setlength\@fpsep{8pt plus 1fil} 
\setlength\@fpbot{0pt}
\makeatother

\begin{table}[ht]
\begin{tabular}{lllllll}
SUMMARY &  &  &  &  &  &  \\ \cline{1-5}
\multicolumn{1}{c}{\textit{Groups}} & \multicolumn{1}{c}{\textit{Count}} & \multicolumn{1}{c}{\textit{Sum}} & \multicolumn{1}{c}{\textit{Average}} & \multicolumn{1}{c}{\textit{Variance}} &  &  \\ \cline{1-5}
Shared & 12706 & 28118 & 2.21297 & 2.38077 &  &  \\
Specialised & 12706 & 28111 & 2.21242 & 2.45665 &  &  \\ \cline{1-5}
 &  &  &  &  &  &  \\
ANOVA &  &  &  &  &  &  \\ \hline
\multicolumn{1}{c}{\textit{Source of Variation}} & \multicolumn{1}{c}{\textit{SS}} & \multicolumn{1}{c}{\textit{df}} & \multicolumn{1}{c}{\textit{MS}} & \multicolumn{1}{c}{\textit{F}} & \multicolumn{1}{c}{\textit{P-value}} & \multicolumn{1}{c}{\textit{F crit}} \\ \hline
Between Groups & 0.00193 & 1 & 0.00193 & 0.00080 &  \textit{\textbf{0.97748}} & 3.84183 \\
Within Groups & 61459.38 & 25410 & 2.41871 &  &  &  \\ \hline
 &  &  &  &  &  & 
\end{tabular}
\caption{Significant test results on the test split of BPIC 2012 W Complete.}
\label{tab:st4}
\end{table}

\subsection*{A.2. Model Interpretations for BPIC~2017 Event Log}

\begin{figure}[htbp]
    \vspace*{-2.5\baselineskip}
    \centering
    \includegraphics[width=0.7\textwidth]{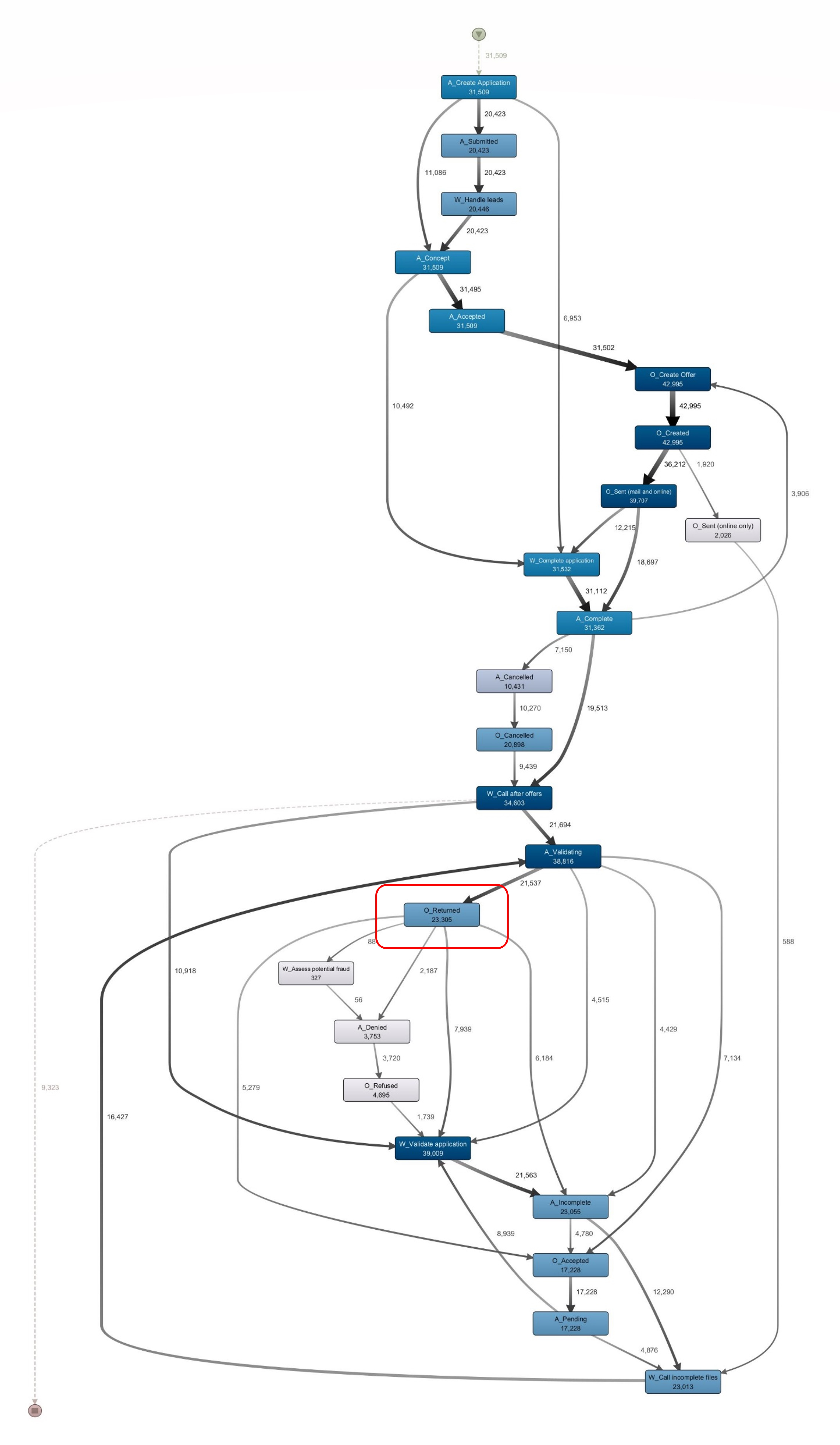}
    \vspace*{-\baselineskip}
    \caption{Process Map of BPIC 2017 Loan Application Process}
    \label{figure:bpic2017_procesmap}
\end{figure}

With the objective of validating our approach, we replicated our models and explanation mechanism on the BPIC 2017 event log~\cite{Dongen2017}. BPIC 2017 event log belongs to an enhanced version of the loan application process that was scrutinised in our main experiments with the BPIC 2012 log. The overall process map generated for the BPIC 2017 log is depicted in Figure~\ref{figure:bpic2017_procesmap}. 

\begin{figure}[htbp]
    \centering
    \includegraphics[width=\textwidth]{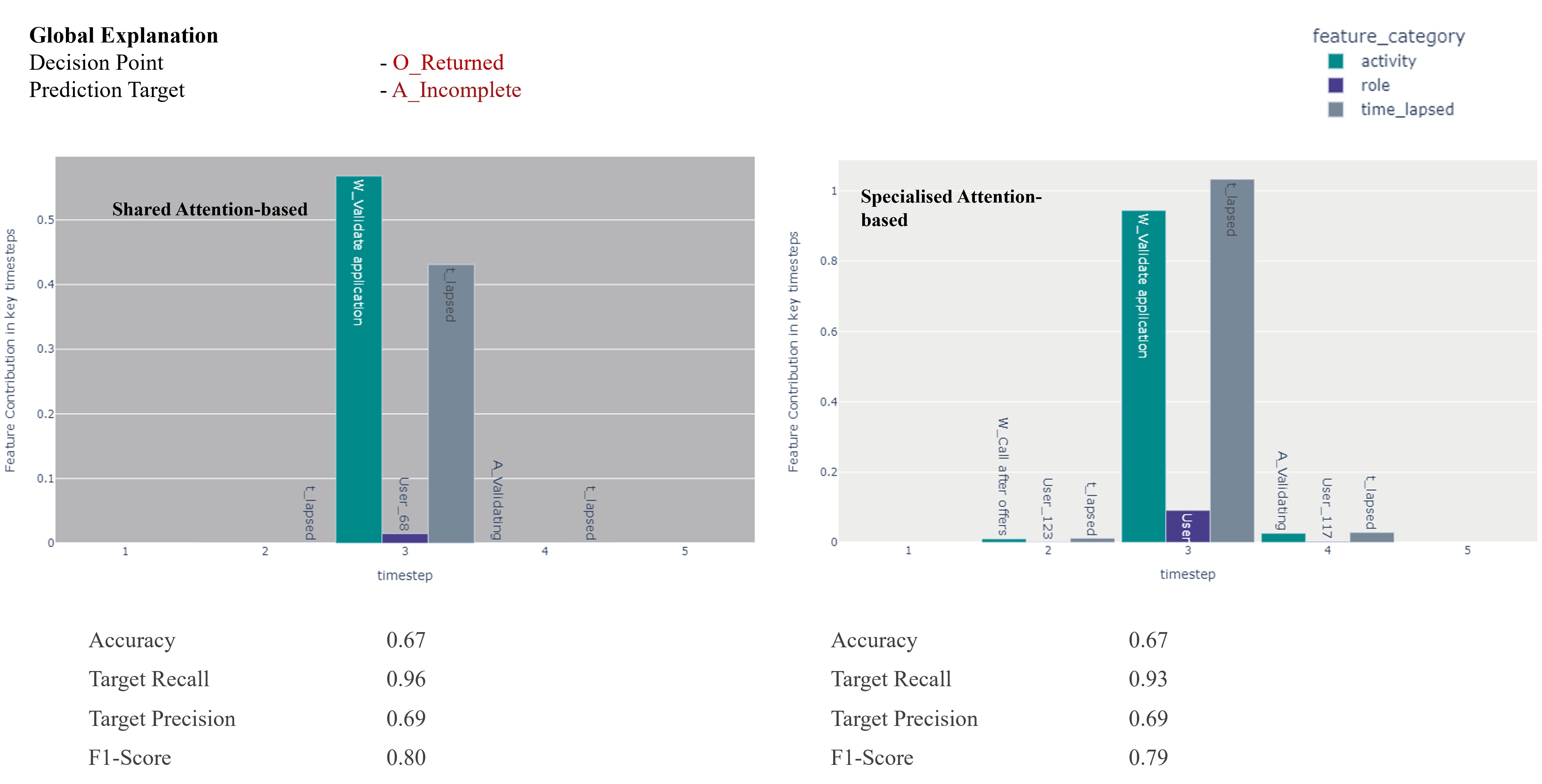}
    \caption{Global Explanation for `A\_Incomplete' target at `O\_Returned' decision point}
    \label{figure:A_Inc_Global}
\end{figure}

\begin{figure}[htbp]
    \centering
    \includegraphics[width=\textwidth]{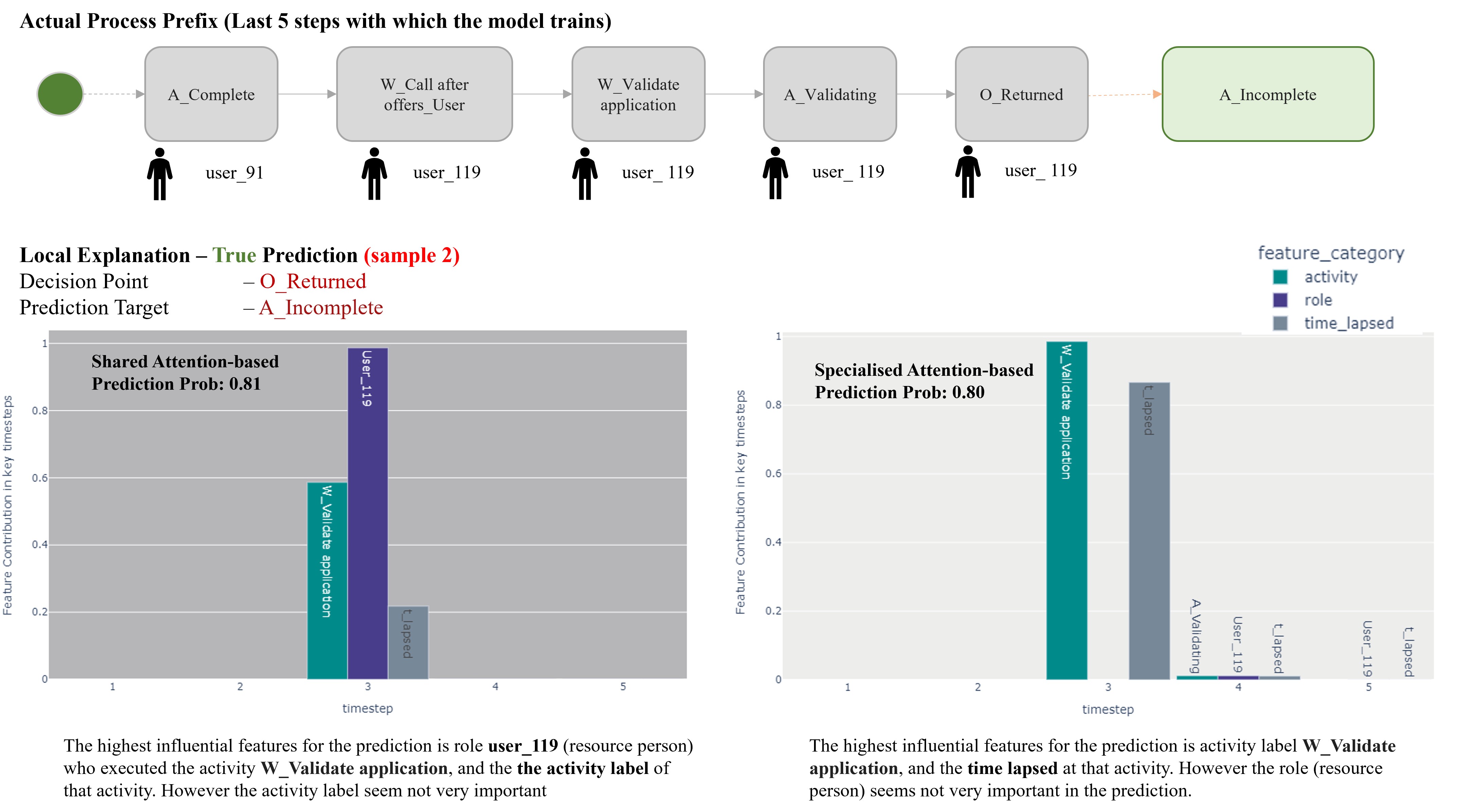}
    \caption{Local Explanation for a True prediction for a `A\_Incomplete' target at `O\_Returned' decision point}
    \label{figure:A_Inc_Local}
\end{figure}

For the Illustration purpose of model explanations, we have chosen the decision point `O\_Returned'  (outlined in red on the process map). Our model predicts the possibility of the process trace moving into either of the next activities `A\_Incomplete', `W\_Validate Applications', or `O\_Accepted', once it passes the decision point `O\_Returned' (see Figures~\ref{figure:A_Inc_Global} and~\ref{figure:A_Inc_Local}). 


\begin{figure}[htbp]
    \centering
    \includegraphics[width=\textwidth]{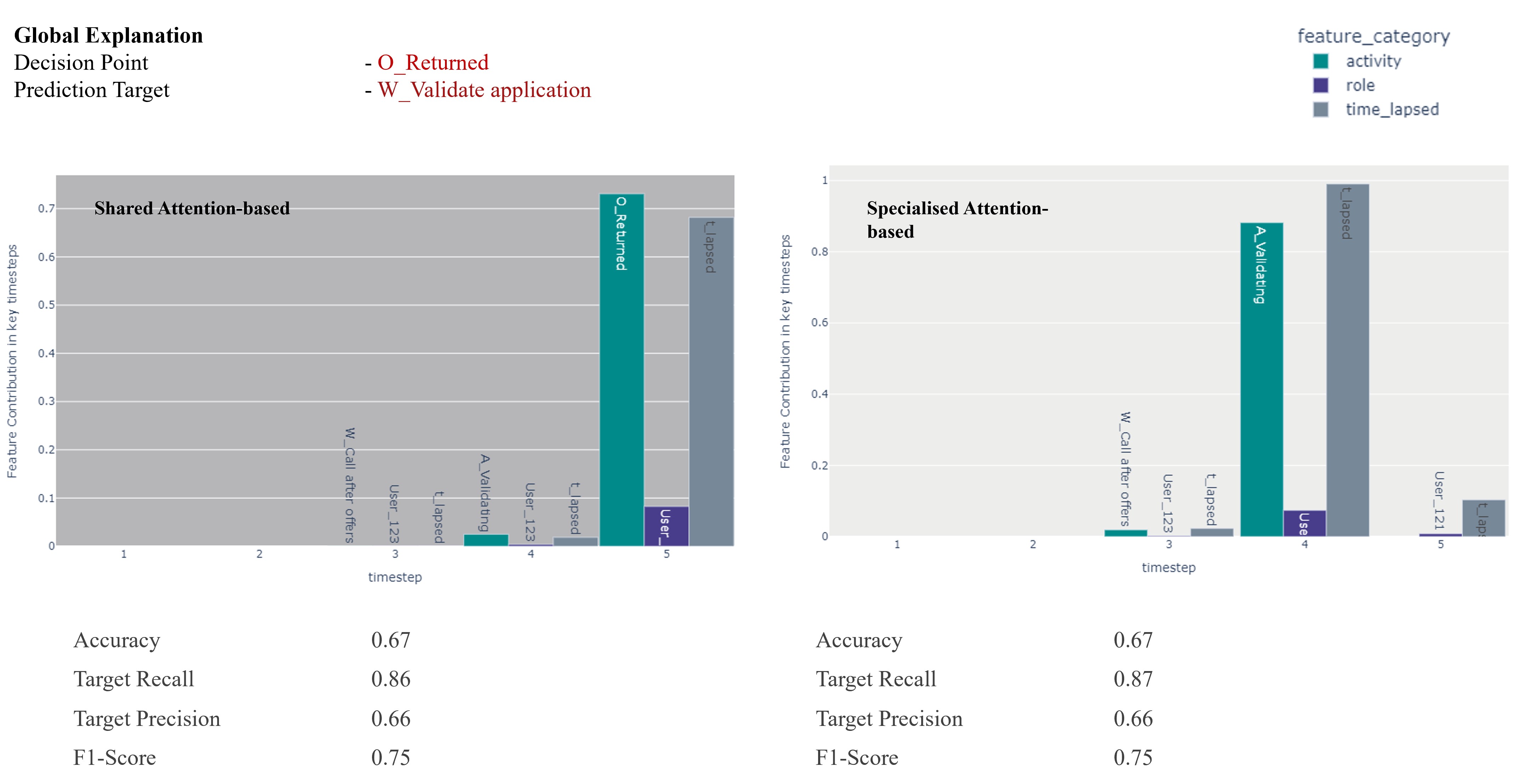}
    \caption{Global Explanation for `W\_Validate Application' target at `O\_Returned' decision point}
    \label{figure:W_Val_Global}
\end{figure}

\begin{figure}[htbp]
    \centering
    \includegraphics[width=\textwidth]{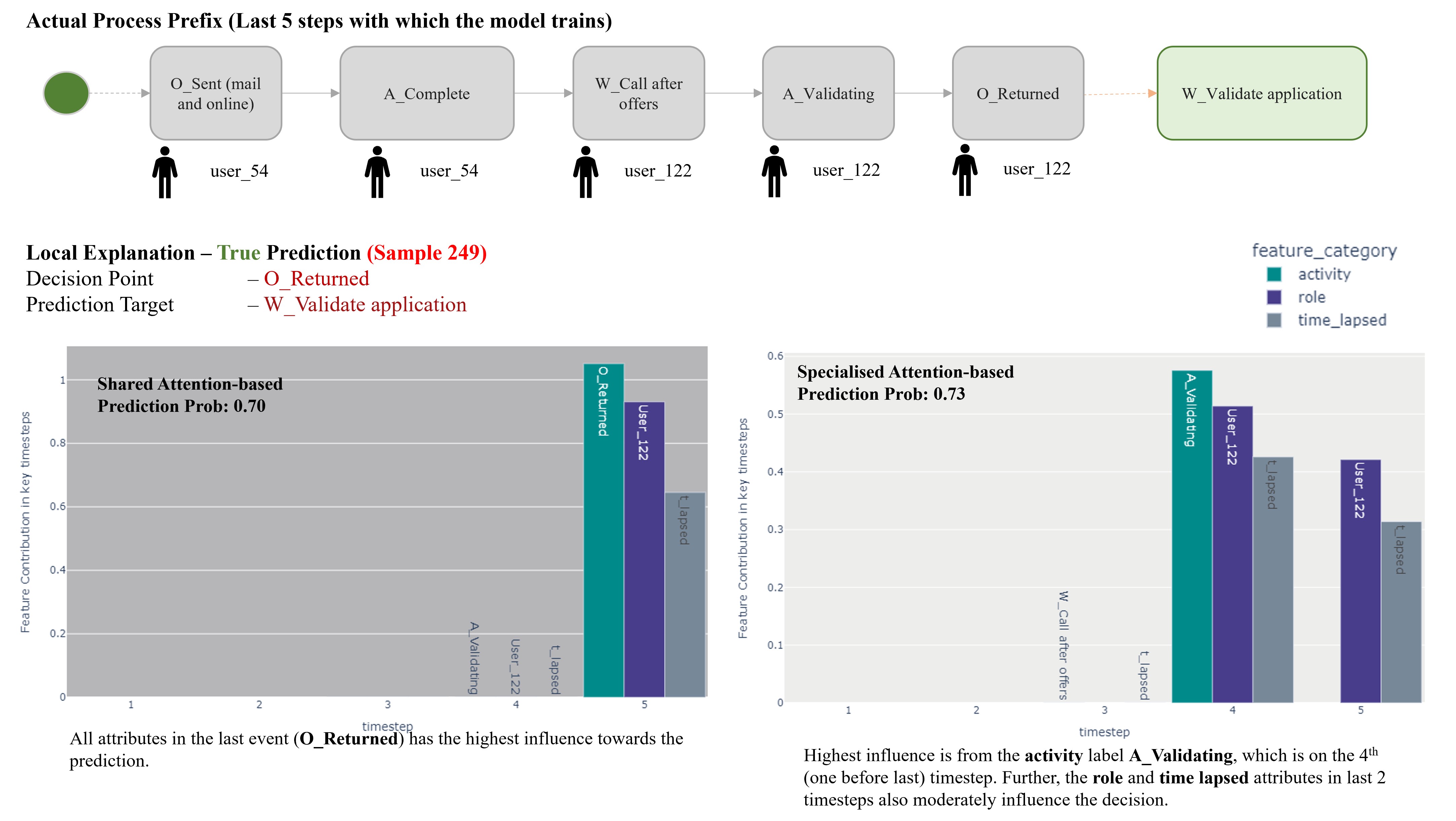}
    \caption{Local Explanation for a True prediction for a `W\_Validate Application' target at `O\_Returned' decision point}
    \label{figure:W_Val_Local}
\end{figure}

For the prediction target of `A\_Incomplete', global interpretation for both models look quite similar. The model primarily pays the attention to the third event of the prefix trace,activity label `W\_Validate Application' and the `time elapsed' at that event. However, in the local interpretation instance analysed, we can observe that the shared attention-based model pays a higher attention to the 'role' attribute at the third event (`role\_119'), whereas the local interpretation of the specialised attention-based model looks quite similar to the global interpretation (see Figures~\ref{figure:W_Val_Global} and~\ref{figure:W_Val_Local}).


\begin{figure}[htbp]
    \centering
    \includegraphics[width=\textwidth]{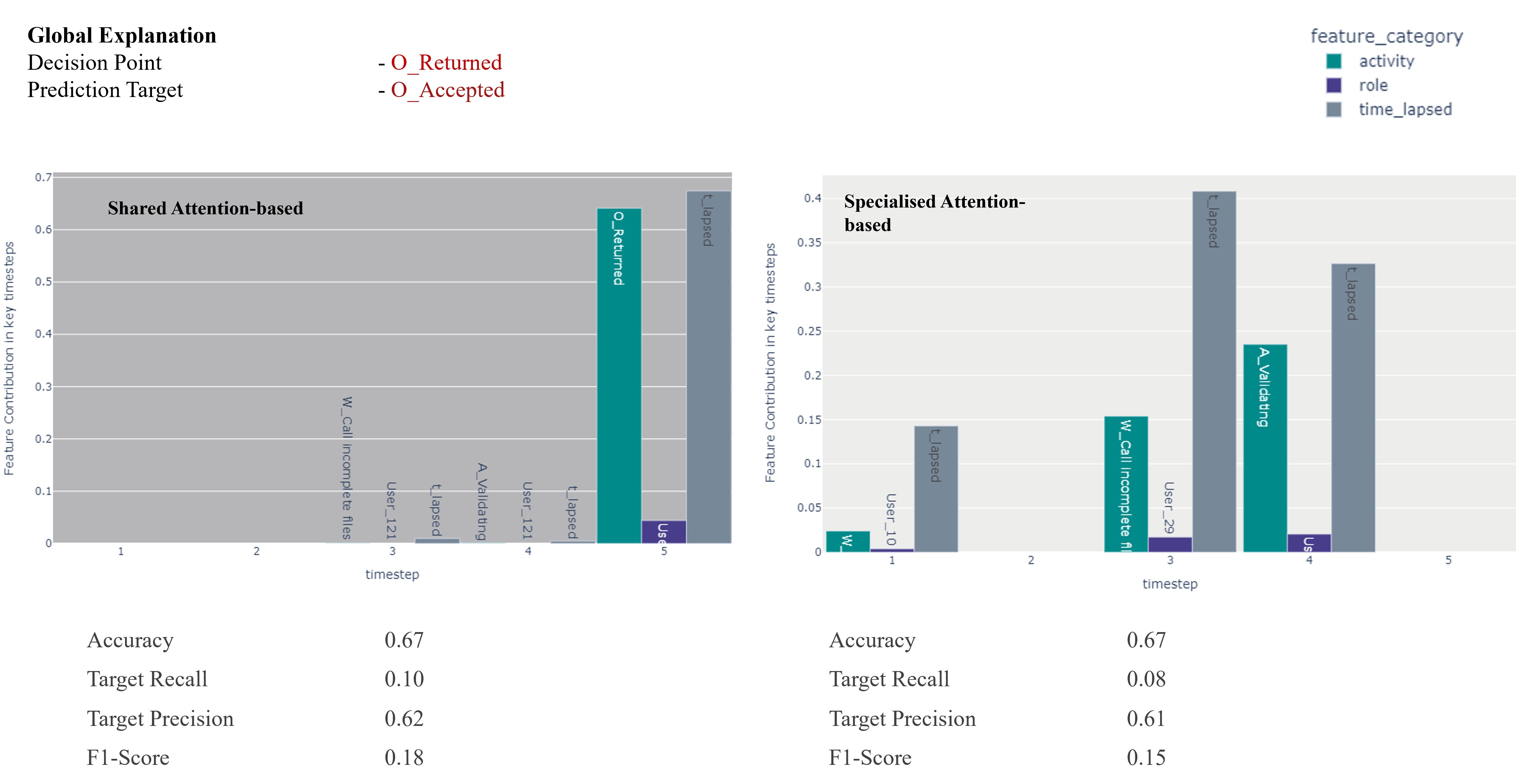}
    \caption{Global Explanation for O\_Accepted' target at `O\_Returned' decision point}
    \label{figure:O_Acc_Global}
\end{figure}

\begin{figure}[htbp]
    \centering
    \includegraphics[width=\textwidth]{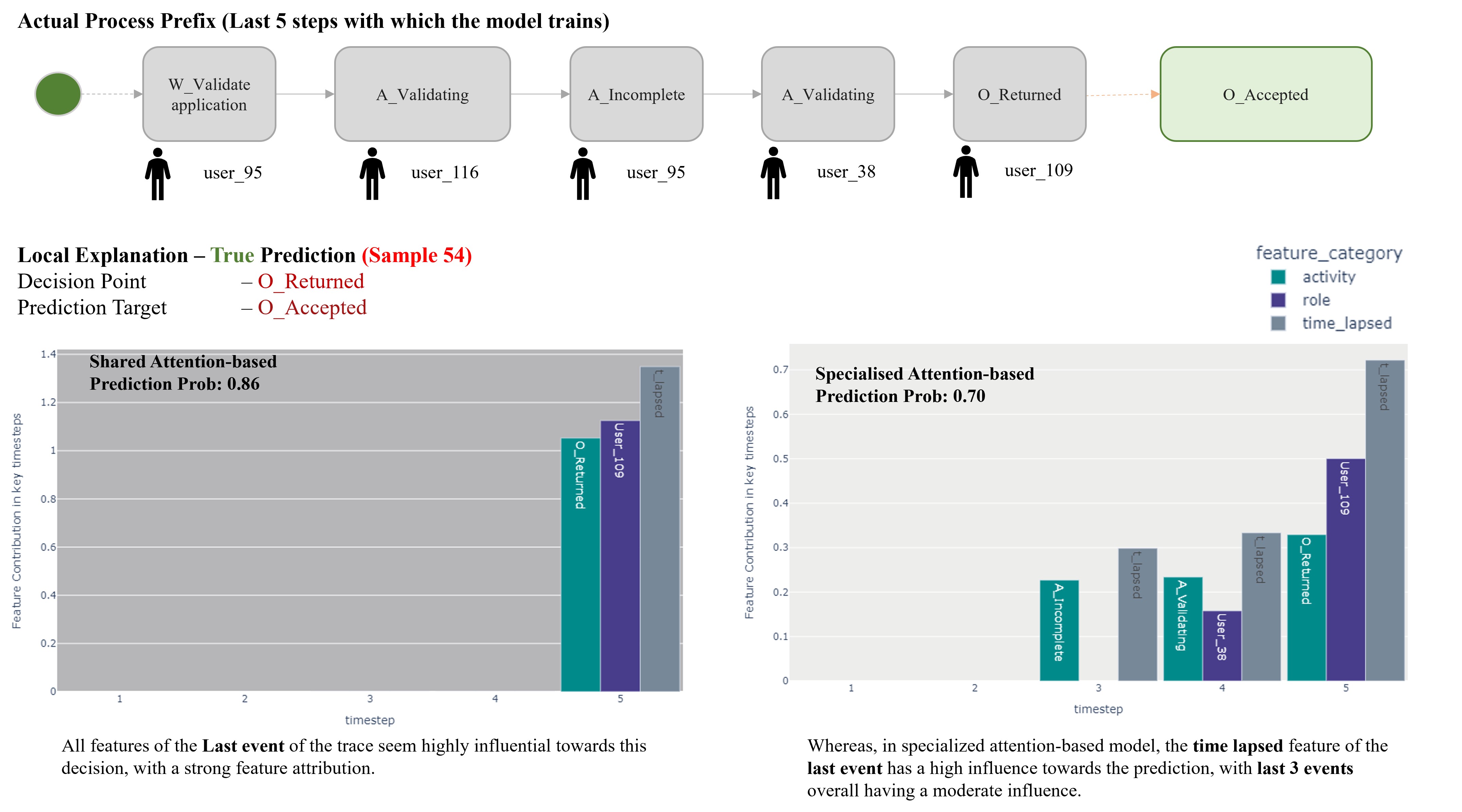}
    \caption{Local Explanation for a True prediction for a `O\_Accepted' target at `O\_Returned' decision point}
    \label{figure:O_Acc_Local_True}
\end{figure}

The global interpretations for the prediction target `W\_Validate Applications are quite contrastive between the two models. As per the shared attention-based mechanism, the most influential feature for the prediction is the activity label of last event of the trace (`O\_Returned', which is the decision point itself) and the `time elapsed' at that point. However, the activity label `O\_Returned' at the last event being common for all the prefixes that goes through the model, a concern arises how valid it is to consider that label to make a particular prediction. As per the specialised attention-based mechanism, the model pays attention to the fourth event of the prefix; the activity label `A\_Validate' and the `time elapsed' feature specifically. The local interpretations of the two models for the same prediction target quite resembles the respective global interpretations, with additionally the 'role' feature category also being influential (see Figures~\ref{figure:O_Acc_Global} and~\ref{figure:O_Acc_Local_True}).


\begin{figure}[htbp]
    \centering
    \includegraphics[width=\textwidth]{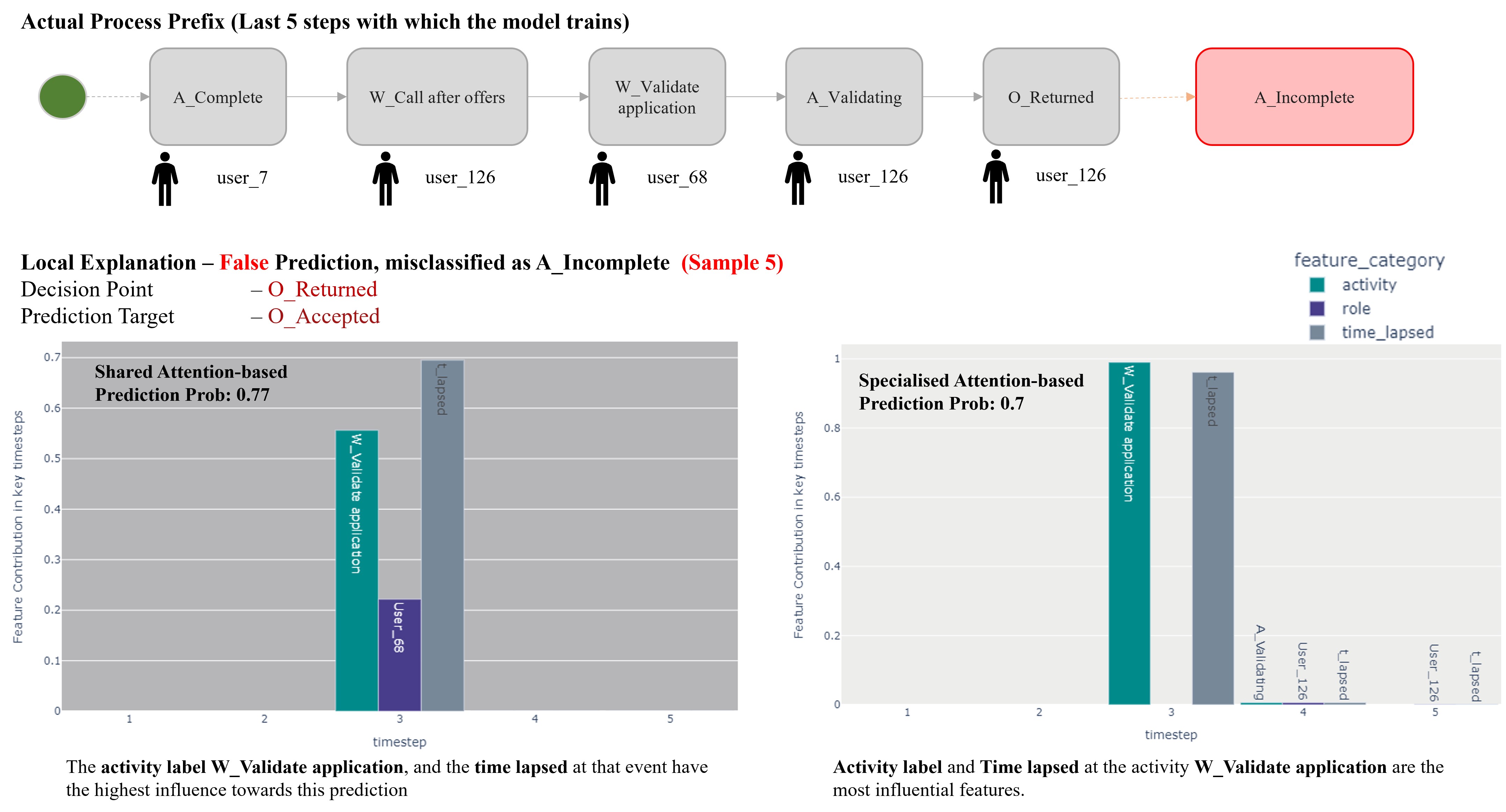}
    \caption{Local Explanation for the misclassified `O\_Accepted' target as a `A\_Incomplete' prediction}
    \label{figure:O_Acc_Local_False_1}
\end{figure}

\begin{figure}[htbp]
    \centering
    \includegraphics[width=\textwidth]{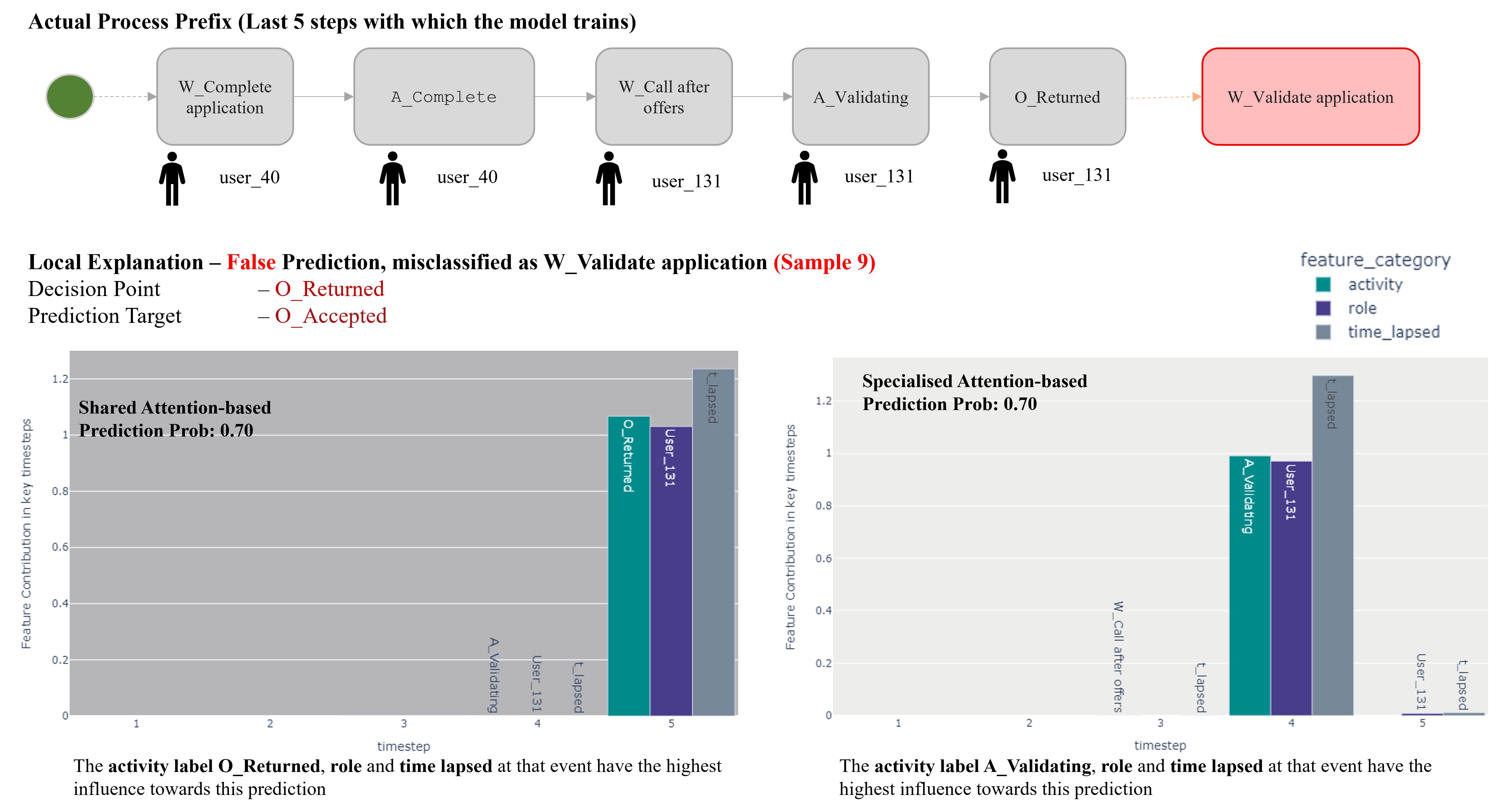}
    \caption{Local Explanation for the misclassified `O\_Accepted' target as a `W\_Validate Application' prediction}
    \label{figure:O_Acc_Local_False_2}
\end{figure}

Unlike the first two prediction targets, the prediction recall of the target `O\_Accepted' is very poor. It looks like it often gets misclassified into either of the `A\_Incomplete' or `W\_Validate Application' targets. When the global interpretations are analysed, we can observe that the shared attention-based model considers the last event (decision point itself) as influential to the decision. Whereas the specialised attention-based model considers the third and fourth events to make decision. For a true prediction, the local interpretation of the shared attention-based model is quite consistent with the global interpretation of the same model. However, the specialized attention-based model seems to pay attention to many features to make a reasonable guess (see Figures~\ref{figure:O_Acc_Local_False_1} and~\ref{figure:O_Acc_Local_False_2}).


Interestingly, the interpretations for two false predictions for the prediction target `O\_Accepted', each got misclassified as either `A\_Incomplete' or `W\_Validate Application' quite closely mimic the global interpretations of the respective targets (see Figure~\ref{figure:O_Acc_Local_False_1} vs Figure~\ref{figure:A_Inc_Global} and Figure~\ref{figure:O_Acc_Local_False_2} vs Figure~\ref{figure:W_Val_Global}).

\color{black}

\end{document}